\documentclass{article}
\usepackage{spconf,amsmath,graphicx}
\usepackage{subfigure}
\usepackage{wrapfig}
\usepackage{xcolor}
\usepackage{mdframed}
\usepackage{colortbl,array}
\usepackage{url}
\usepackage{float}

\definecolor{verylightgray}{gray}{0.9}

\newcommand{\gray}[2][0.4]{\textcolor[gray]{#1}{#2}}

\newmdenv[
  backgroundcolor=verylightgray, % Set the background color
  skipabove=\topsep,
  skipbelow=\topsep,
  leftmargin=3mm,
  rightmargin=3mm,
  innertopmargin=1mm,  %\baselineskip,
  innerbottommargin=1mm,  %\baselineskip,
  innerleftmargin=1mm,
  innerrightmargin=1mm
]{coloredquotation}

\title{Iterative Self-Improvement of Vision Language Models \\for Image Scoring and Self-Explanation}
\twoauthors
{Naoto Tanji\sthanks{Email address: \texttt{naoto.tanji@septeni.co.jp}}}
	{Septeni Japan, Inc.}
{Toshihiko Yamasaki}
	{The University of Tokyo}
\begin{document}
%\ninept
%
\maketitle
\begin{abstract}
Image scoring is a crucial task in numerous real-world applications. To trust a model's judgment, understanding its rationale is essential. This paper proposes a novel training method for Vision Language Models (VLMs) to generate not only image scores but also corresponding justifications in natural language. Leveraging only an image scoring dataset and an instruction-tuned VLM, our method enables self-training, utilizing the VLM's generated text without relying on external data or models. In addition, we introduce a simple method for creating a dataset designed to improve alignment between predicted scores and their textual justifications. By iteratively training the model with Direct Preference Optimization on two distinct datasets and merging them, we can improve both scoring accuracy and the coherence of generated explanations.
\end{abstract}
\section{Introduction}
\label{sec:intro}
Deep learning is revolutionizing image analysis, enabling automated classification and scoring with enhanced accuracy and efficiency.
Examples include disease detection in medical images, defect identification in quality control, and predicting advertising effectiveness.
For trustworthy applications of deep learning models in these scenarios, the explainability of the model's decisions is crucial.
Numerous methods for explainable Artificial Intelligence (AI) have been investigated \cite{das2020opportunities}.
With recent advances in large language models (LLMs), explanations provided in natural language are superior in terms of human comprehensibility. 
While it is possible for LLMs to provide post-hoc explanations for the decisions made by other models, having the LLM generate its reasoning and thought process alongside its predictions can lead to a more coherent and consistent approach. This integration enhances the overall interpretability of the model outputs, making it easier for users to understand the underlying rationale behind the decisions.

\begin{figure}[t]
\centering
\includegraphics[width=8.8cm]{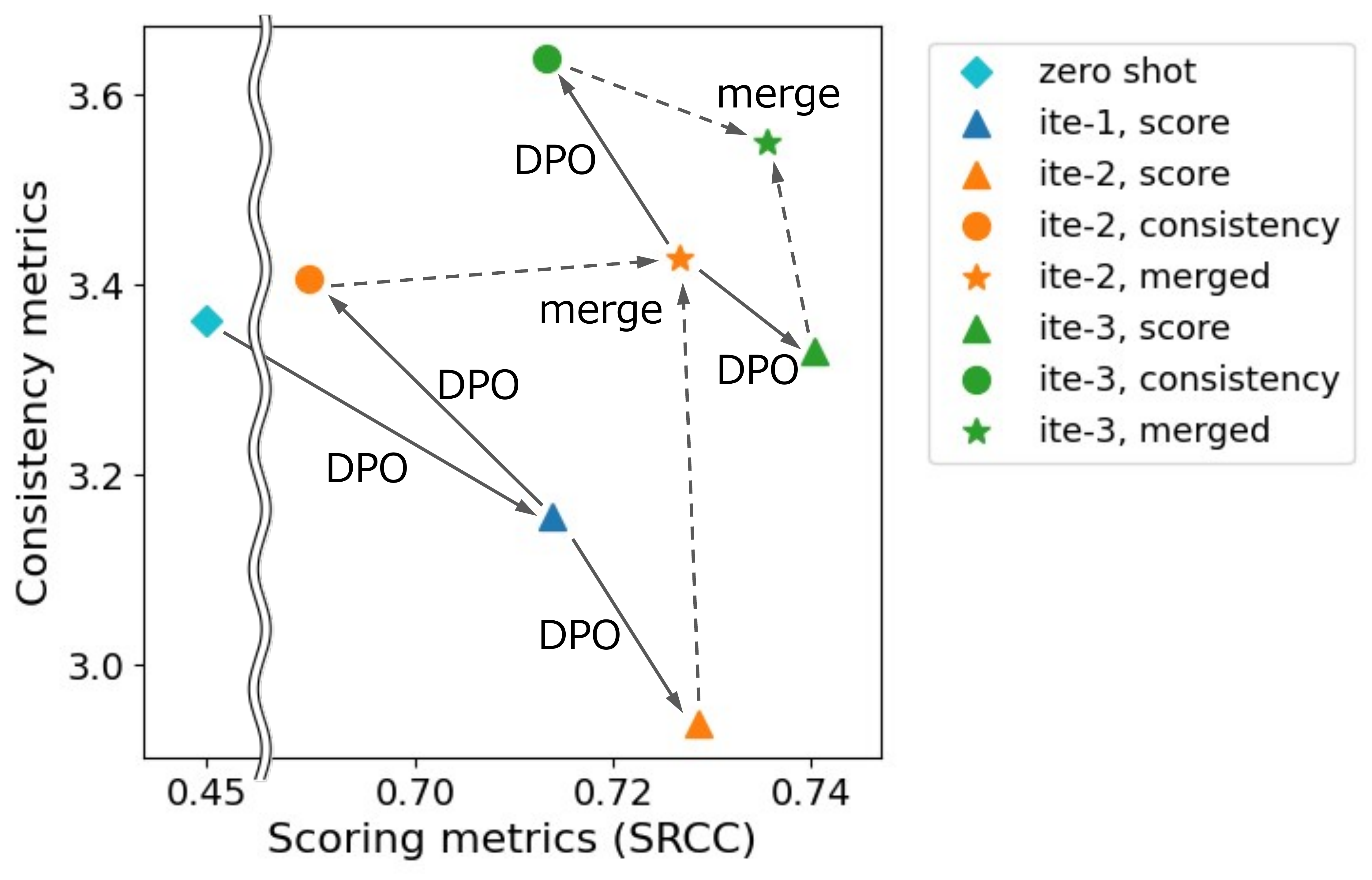}
\vspace{-20pt}
\caption{Illustration of our main result. Iterative training with two types of self-generated data and model merging improves both score prediction accuracy and consistency between explanatory texts and predicted scores.}
\label{fig:main_result}
\end{figure}

In this study, we train VLMs so that they can simultaneously predict scores and provide explanations that justify their decision. A high-performing LLM-based VLM can generate plausible and grammatically correct descriptions that are relevant to the image content, even in a zero-shot setting. However, if the model has not learned the scores provided by the dataset, the credibility of these descriptions may be low. On the other hand, training the model requires providing both scores and corresponding explanations, but obtaining ground truth for these explanations can be challenging. One potential solution is to have humans create the explanations, but this approach is labor-intensive, making it impractical to gather large-scale training data, especially in the context of image scoring.

\begin{figure*}[t]
\centering
\includegraphics[width=11.1cm]{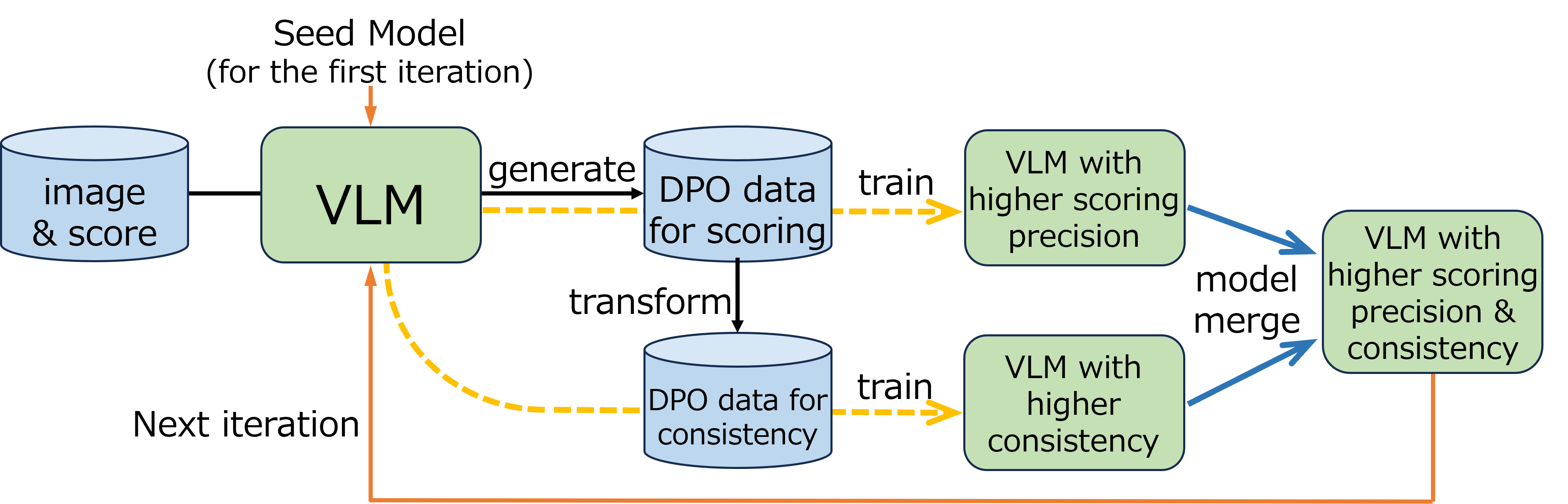}
\vspace{-10pt}
\caption{Overview of our iterative approach: (1) A DPO dataset for score learning is created from an image-score dataset using a VLM. (2) This dataset is post-processed to create a dataset for enhancing consistency between texts and predicted scores. (3) The VLM is trained on both datasets, yielding two specialized models. (4) These models are merged to combine their capabilities.}
\label{fig:overview}
\end{figure*}

To address this issue, we adopt a self-training approach in which the VLM itself generates training data for explanations. By combining the information from a dataset consisting of images and scores with the VLM's capabilities acquired through pretraining and instruction tuning, we can enhance both the score prediction and explanation abilities of the VLM without relying on other models or datasets. 
First, we generate explanations conditioned on the ground truth score, allowing the VLM to create its own training data comprised of explanations. 
Separately, we also generate explanations conditioned on incorrect scores. 
Then, we combine these correct and incorrect score-conditioned explanations to create the training data for Direct Preference Optimization (DPO)~\cite{rafailov2023direct}.
This data is designed to train the model by contrasting correct and incorrect scores. 
Furthermore, by applying a simple post-processing to the DPO dataset used for score prediction training, we create a dataset that aims to improve the consistency between the predicted scores and the content of the explanations. Merging the models trained on these two datasets allows us to develop a model that enhances both score prediction accuracy and the coherence of the explanations. As illustrated in Figure 1, iterating this process effectively improves both capabilities.

\section{Related Work}
\label{sec:related}

\noindent
\textbf{Image Aesthetic Assessment} \hspace{3pt}
As an example of a high-level task where natural language explanations can be particularly valuable, we tackle image aesthetic assessment (IAA) in this work.
IAA datasets (e.g., AVA~\cite{murray2012ava} and AADB~\cite{kong2016aesthetics}) consist of photographs scored by human annotators.
Because IAA relies on subjective human perception, generating human-understandable natural language explanations for aesthetic judgments can lead to more trustworthy and user-friendly models.

While extensive research exists on IAA~\cite{deng2017image}, most studies focus solely on predicting aesthetic scores without providing explanations.
Some recent work has applied VLMs to IAA, such as VILA~\cite{ke2023vila} and Q-ALIGN~\cite{wu2024qalign}. 
VILA pretrains a VLM on image-comment pairs, achieving strong performance, including zero-shot, on aesthetic prediction and image style classification. While capable of generating image captions due to its pretraining objective, VILA doesn't generate explanations for its aesthetic judgments. Furthermore, our work differs by not relying on any external text data for training.
Q-ALIGN reframes score prediction as a text generation task, leveraging pretrained VLMs to achieve state-of-the-art performance. Similar to Q-ALIGN, we also treat score prediction as a text generation task. 
However, our focus is on leveraging the text generation capabilities of VLMs to produce explanations for the aesthetic judgments.

\noindent
\textbf{Natural language explanations by LLM} \hspace{3pt}
Recent advancements in the field of natural language processing 
allow models to explain their reasoning in natural language. 
Chain-of-Thought (CoT) reasoning enhances the model's reasoning abilities by prompting the model to generate step-by-step reasoning~\cite{wei2022chain}.
For logical tasks, these reasoning processes enhance the interpretability of the model behavior. 
However, the primary goal of CoT reasoning is to improve the accuracy of the model's predictions, not to explain its justification.
The latter is the focus of our research.

Similar to our work, there are lines of research that have focused on explaining model predictions using natural language. Some approaches employ supervised learning with human-written explanations~\cite{park2018multimodal,narang2020wt5,wiegreffe2021measuring},
while others leverage few-shot learning techniques~\cite{marasovic2022few,wiegreffe2022reframing}.
In contrast, our method enables the target model itself to generate its own explanation training data and to enhance the quality of explanations in a bootstrapping way.
This eliminates the need for human-generated data or data generated by an external model, which is a key advantage.

\noindent
\textbf{Self-improvement of LLM} \hspace{3pt}
Self-improvement approaches for LLMs, where feedback through reflection or self-generated training data enhances performance, have seen a surge in research recently~\cite{huang2023large,madaan2023self,yuan2024self,pang2024iterative}.
Particularly relevant to our work is~\cite{pang2024iterative}, which has an LLM generate multiple CoT reasonings and predictions, and then selects correct and incorrect predictions to create DPO data. 
Our approach shares similarities, but by conditioning the explanation generation on the prediction, we avoid the need for multiple generations and post-selection, reducing the computational overhead. 
In the vision-language domain,~\cite{deng2024enhancing} demonstrates the effectiveness of self-generated DPO datasets, successfully improving image comprehension capabilities through self-training.

\begin{figure}[t]
\centering
\includegraphics[width=8.5cm]{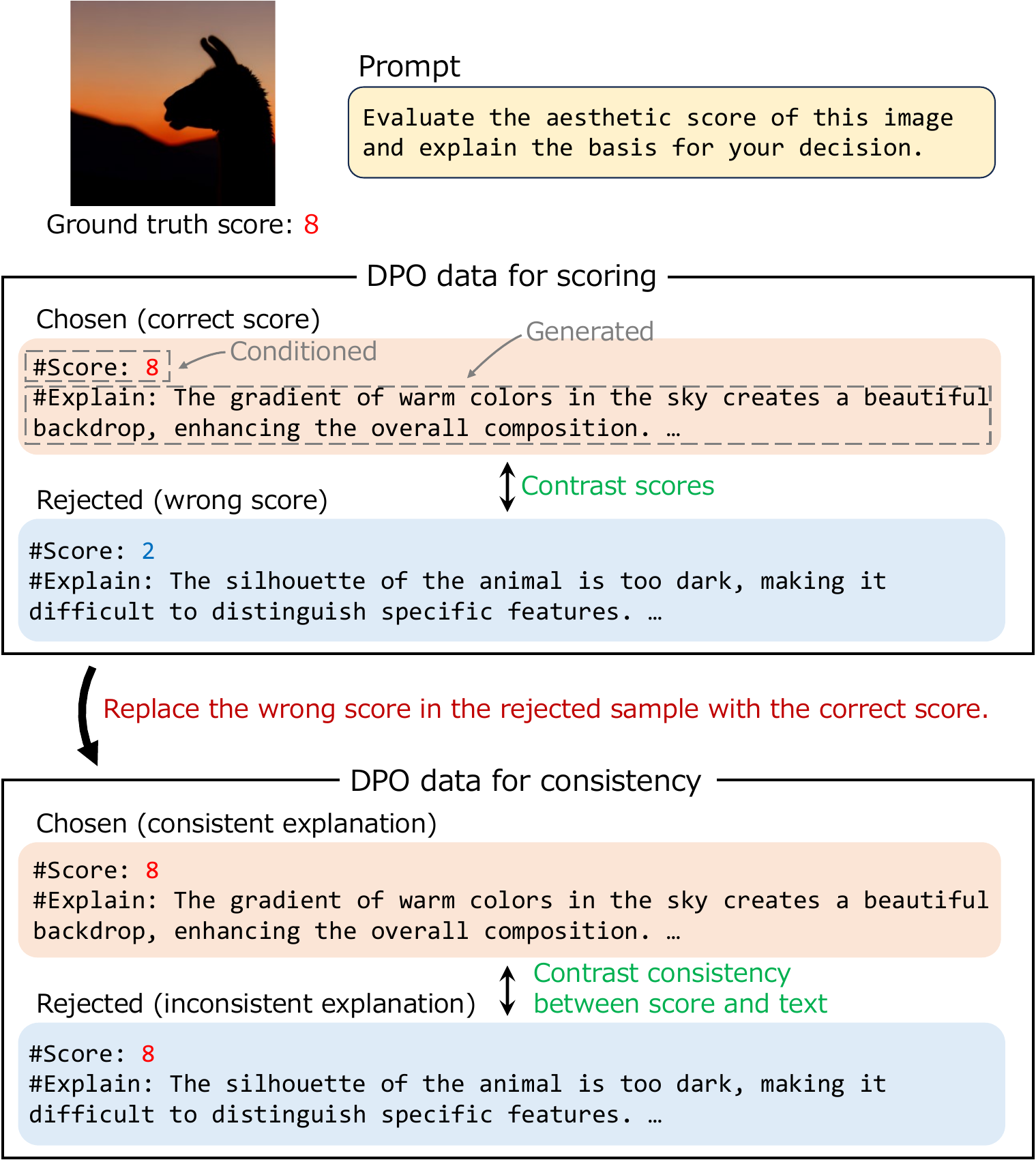}
\vspace{-10pt}
\caption{Illustration of our datasets.
DPO data for learning score prediction consists of ``chosen" and ``rejected" samples, which are conditioned on the ground truth score and an incorrect score, respectively. 
To further enhance score-explanation consistency, we create additional DPO data by replacing the incorrect scores in the rejected samples with the correct scores, resulting in pairs that contrast the consistency between score and explanatory text.
The prompt is provided for illustrative purposes.}
\label{fig:data}
\end{figure}

\section{Method}
\label{sec:method}
The overview of our method is illustrated in Fig.~\ref{fig:overview}.
We start from an instruction-tuned VLM and a target dataset that consists of images and scores. 
The first step is generating two kinds of DPO datasets; one is for learning the score prediction task, and the other is for improving the alignment between the predicted score and its corresponding explanations. 
Using these datasets, we train the VLM to obtain two specialized models; one has higher accuracy for the score prediction, and the other can generate explanations that are more aligned with the predicted scores. Then, we merge the two models to unify their distinct abilities. 
These processes can be iterated to gradually enhance the model's abilities.

\subsection{Data Generation}
\label{ssec:data_generation}
Our data generation strategy is presented in Fig.~\ref{fig:data}. 
We use a prompt that instructs the VLM to output a score first and then explanation texts that justify the decision\footnote{For actual prompts and other details, see Appendix~\ref{sec:details}.}.
We choose the predict-then-explain strategy since it enables us to intervene in the VLM's generation process by injecting an arbitrary score into the prompt. 
After the prompt, we append the ground truth score to it, and let the VLM generate the continuation of the text, namely an explanation that justifies the assigned score as if the score is predicted by the VLM itself.
The texts generated by this process are employed as the ``chosen response" in the DPO dataset. 
For ``rejected response", we do the same thing but with an incorrect score, which is chosen randomly while excluding values close to the correct score.
The VLM can learn the score prediction capabilities from this DPO dataset by contrasting the ground truth score and an incorrect score. 
Because this DPO dataset includes explanation texts, the VLM can retain its language ability during the training process, unlike when it is trained solely on scores.

An issue concerning the LLM's self-explanation of its decision is whether the explanation is consistent with the decision. 
We quantitatively verified that the descriptions in the DPO dataset tend to align with the scores to a certain extent by using LLM-as-a-judge. 
This is due to the coherent text generation capabilities acquired by the VLM through pretraining and instruction tuning. 
To further enhance this alignment, we create a new DPO dataset. 
The method is simple: we retrospectively replace the incorrect scores assigned to the rejected samples with the ground truth scores. 
After this replacement, both chosen and rejected samples are assigned the same ground truth scores, but their explanation texts differ. 
The chosen samples are texts generated by the VLM based on the ground truth scores, while the rejected samples are based on incorrect scores. 
The rejected samples after the replacement are more likely to have explanations that do \textit{not} align with the ground truth scores. 
By contrasting the chosen and rejected samples in this dataset during DPO training, we can improve the alignment between scores and explanations.

\subsection{Training and Model Merge}
\label{ssec:training_merge}
We iteratively self-train the VLM to improve its abilities for score prediction and consistent explainability. 
Following the data generation recipe introduced above, the DPO datasets are generated by the model in the previous iteration.

In the first iteration, since the initial model's score prediction ability is very low, we train the model only on the DPO dataset for the score prediction in order to first enhance the score prediction capability.
In the second and later iterations, we use both datasets simultaneously for training.
A simple way to train a model on two datasets is to merge the separately trained models. We first train individual models on each dataset and then combine these models into a single one.
For the merging method, we adopt TIES-Merging~\cite{yadav2023ties}.

\section{Experimental Setup}
\label{sec:setup}

\subsection{Datasets}
\label{ssec:datasets}
As IAA datasets, we use AVA~\cite{murray2012ava} and AADB~\cite{kong2016aesthetics}. 
We follow the train-val-test split given by~\cite{talebi2018nima} for AVA, and the official one for AADB. 
In these datasets, each image is rated by several users with integer scores. We use average scores as ground truth labels. 

To treat score prediction as a language generation task, we discretize the original scores into ten bins based on the quantiles of the training data distribution and assign integer scores from 0 to 9.

\subsection{Models and training}
\label{ssec:models}
We experimented with several VLMs, ranging in size from 0.5 billion to 7 billion parameters:  LLaVA-interleave-0.5B~\cite{li2024llava}, InternVL2-2B~\cite{chen2024internvl}, LLaVA-1.5-7B~\cite{liu2024improved}, and LLaVA-NeXT-7B~\cite{liu2024llavanext}.
Due to limited computational resources, we fine-tuned the models using Low-Rank Adaptation (LoRA)~\cite{hu2022lora}. 
LoRA adapters were applied to all linear layers within the language model component. For DPO, the base model without the LoRA adapter served as the reference model. 

\subsection{Evaluation}
\label{ssec:evaluation}
Our proposed training method aims to improve both score prediction accuracy and the consistency between the predicted score and its explanation. Therefore, we adopt evaluation metrics that capture both score accuracy and explanation consistency. 

For score prediction, we use Pearson's linear correlation coefficient (PLCC) and Spearman's rank correlation coefficient (SRCC) between the predicted and ground truth scores. These metrics are computed for the original scores given by the datasets. To convert the integer scores predicted by the model back to the original scores, we employ a method similar to the one introduced in~\cite{wu2024qalign}.
We compute the probability distribution of the integer scores by $p_i = \exp(l_i) / \sum_j \exp(l_j)$, where $l_i$ is the logit for the token corresponding to the score $i$ ($=0,1,\cdots,9$). 
The predicted score in the original scale is then given by $\sum_i \bar{s}_i p_i$, where $\bar{s}_i$ represents the reference value in the original scale for bin $i$.
We optimize $\bar{s}_i$ by the least squares fit to validation data after training the model.

To assess the quality of the explanations, we use LLM-based evaluation as a proxy for human evaluation. 
We input the predicted score and its explanation into GPT-4o (\texttt{{\footnotesize gpt-4o-2024-08-06}}) and let it assess the degree of consistency between the predicted score and the accompanying explanation.
We convert the 5-point scale to integer scores ranging from 0 to 4, and report the average value over the test set as consistency score (abbreviated as \texttt{{\small Cons}}). 
Note that the external model is used only for the post-evaluation. The VLM is trained only by data generated by itself.

\section{Results}
\label{sec:results}

\begin{table}[t]
  \caption{Results of our method applied to the LLaVA-NeXT-7B model on the AVA test dataset.
  For reference, we also include the accuracy of some previous studies (* indicates a different train-test split from ours).
  }
  \label{table:main_result}
  \vspace{3pt}
  \centering
  \setlength{\tabcolsep}{3pt}
  {\small 
  \begin{tabular}{rr|rr|r}
    \hline
    {} & {} & PLCC &  SRCC & Cons \\
    \hline \hline
    {} & zero-shot & 0.452 & 0.446& 3.36 \\
    \hline
    ite-1 & score & 0.719 & 0.714& 3.16 \\ 
    \hline
    {} & score & 0.722 & 0.729 & 2.94 \\[-2pt]
    {} & consistency & 0.688 & 0.689 & 3.41 \\[-2pt] 
    ite-2 & merged & 0.723 & 0.727 & 3.43 \\
    \hline
    {} & score & 0.738 & 0.740 & 3.33 \\[-2pt]
    {} & consistency & 0.676 & 0.713 & 3.64 \\[-2pt] 
    ite-3 & merged & 0.722 & 0.735 & 3.55 \\
    \hline
    {} & score & 0.741 & 0.745 & 3.50 \\[-2pt]
    {} & consistency & 0.618 & 0.710 & 3.64 \\[-2pt] 
    ite-4 & merged & 0.716 & 0.739 & 3.57 \\
    \hline \hline
    {} & \gray{NIMA~\cite{talebi2018nima}} & \gray{0.636} & \gray{0.612} & \gray{-} \\
    {} & \gray{MUSIQ~\cite{ke2021musiq}*} & \gray{0.738} & \gray{0.726} & \gray{-} \\
    {} & \gray{VILA~\cite{ke2023vila}*} & \gray{0.774} & \gray{0.774} & \gray{-} \\
    {} & \gray{Q-ALIGN~\cite{wu2024qalign}*} & \gray{0.817} & \gray{0.822} & \gray{-} \\
    \hline
  \end{tabular}
  }
\end{table}

\subsection{Main Results}
\label{ssec:main_results}
Table~\ref{table:main_result} presents the results of our method applied to the LLaVA-NeXT-7B model on the AVA dataset, a part of which is also plotted in Fig.~\ref{fig:main_result}. 
Zero-shot refers to the initial model that is not yet trained by our method, which results in low accuracies for the score prediction, since the model tends to predict scores clustered around the middle.
By training the model with the DPO dataset for scoring in the first iteration, the scoring metrics are significantly improved, while the consistency score is degraded.
From the second iteration, we train the model with the two distinct datasets and observe that the scoring metrics and the consistency metrics are raised by each dataset, respectively.
Merging these two models yields a model with a balance of their respective capabilities.
For this model and dataset, we see that the training effects are saturated at the fourth iteration. 

For reference, the accuracy of several previous studies on the AVA dataset is also shown in Table~\ref{fig:main_result}. While our primary goal is to enable the model to explain its reasoning, and not necessarily to maximize prediction accuracy, our approach achieves accuracy comparable to, or even exceeding, that of representative CNN- and Vision Transformer-based models~\cite{talebi2018nima,ke2021musiq}. It does not, however, outperform state-of-the-art VLM-based methods~\cite{ke2023vila,wu2024qalign}.

\begin{table}[t]
  \caption{Results of all models.
  For iterations 2-4, only the results of the merged models are shown.
  }
  \label{table:all_results}
  \centering
  \begin{minipage}[t]{0.48\textwidth}
  \centering
  (a) AVA dataset
  \setlength{\tabcolsep}{3pt}
  {\small 
  \begin{tabular}{r|r|rrrrr}
    \hline
    {} & {} & zero shot & ite-1 & ite-2 & ite-3 & ite-4 \\
    \hline \hline
    LLaVA- & SRCC & 0.446 & 0.714 & 0.727 & 0.735 & 0.739 \\[-3pt]
    Next-7B & Cons & 3.36 & 3.16 & 3.43 & 3.55 & 3.57 \\[-1pt]
    \hline
    LLaVA- & SRCC & 0.291 & 0.694 & 0.694 & 0.708 & 0.715 \\[-3pt]
    1.5-7B & Cons & 3.20 & 3.15 & 3.35 & 3.47 & 3.47 \\[-1pt]
    \hline
    InternVL2- & SRCC & 0.315 & 0.675 & 0.682 & 0.689 & 0.693 \\[-3pt]
    2B & Cons & 3.28 & 2.87 & 3.39 & 3.29 & 3.37 \\[-1pt]
    \hline
    LLaVA-inter- & SRCC & 0.366 & 0.672 & 0.677 & 0.679 & 0.685 \\[-3pt]
    leave-0.5B & Cons & 2.43 & 2.33 & 2.50 & 2.57 & 2.72 \\[-1pt]
    \hline
  \end{tabular}
  }
  \end{minipage}
  \begin{minipage}[t]{0.48\textwidth}  % To adjust space
  {}
  \end{minipage}
  \begin{minipage}[t]{0.48\textwidth}
  \centering
  (b) AADB dataset
  \setlength{\tabcolsep}{3pt}
  {\small 
  \begin{tabular}{r|r|rrrrr}
    \hline
    {} & {} & zero shot & ite-1 & ite-2 & ite-3 & ite-4 \\
    \hline \hline
    LLaVA- & SRCC & 0.509 & 0.623 & 0.638 & 0.641 & 0.645 \\[-3pt]
    Next-7B & Cons & 3.47 & 3.54 & 3.58 & 3.54 & 3.48 \\[-1pt]
    \hline
    LLaVA- & SRCC & 0.401 & 0.516 & 0.550 & 0.569 & 0.580 \\[-3pt]
    1.5-7B & Cons & 3.05 & 3.05 & 3.32 & 3.36 & 3.39 \\[-1pt]
    \hline
    InternVL2- & SRCC & 0.408 & 0.641 & 0.642 & 0.649 & 0.643 \\[-3pt]
    2B & Cons & 3.23 & 3.19 & 3.65 & 3.62 & 3.57 \\[-1pt]
    \hline 
    LLaVA-inter- & SRCC & 0.218 & 0.568 & 0.587 & 0.592 & 0.590 \\[-3pt]
    leave-0.5B & Cons & 2.33 & 1.36 & 2.00 & 2.18 & 2.16 \\[-1pt]
    \hline
  \end{tabular}
  }
  \end{minipage}
\end{table}

\begin{figure}[t]
  \begin{center}
   \includegraphics[width=8.8cm]{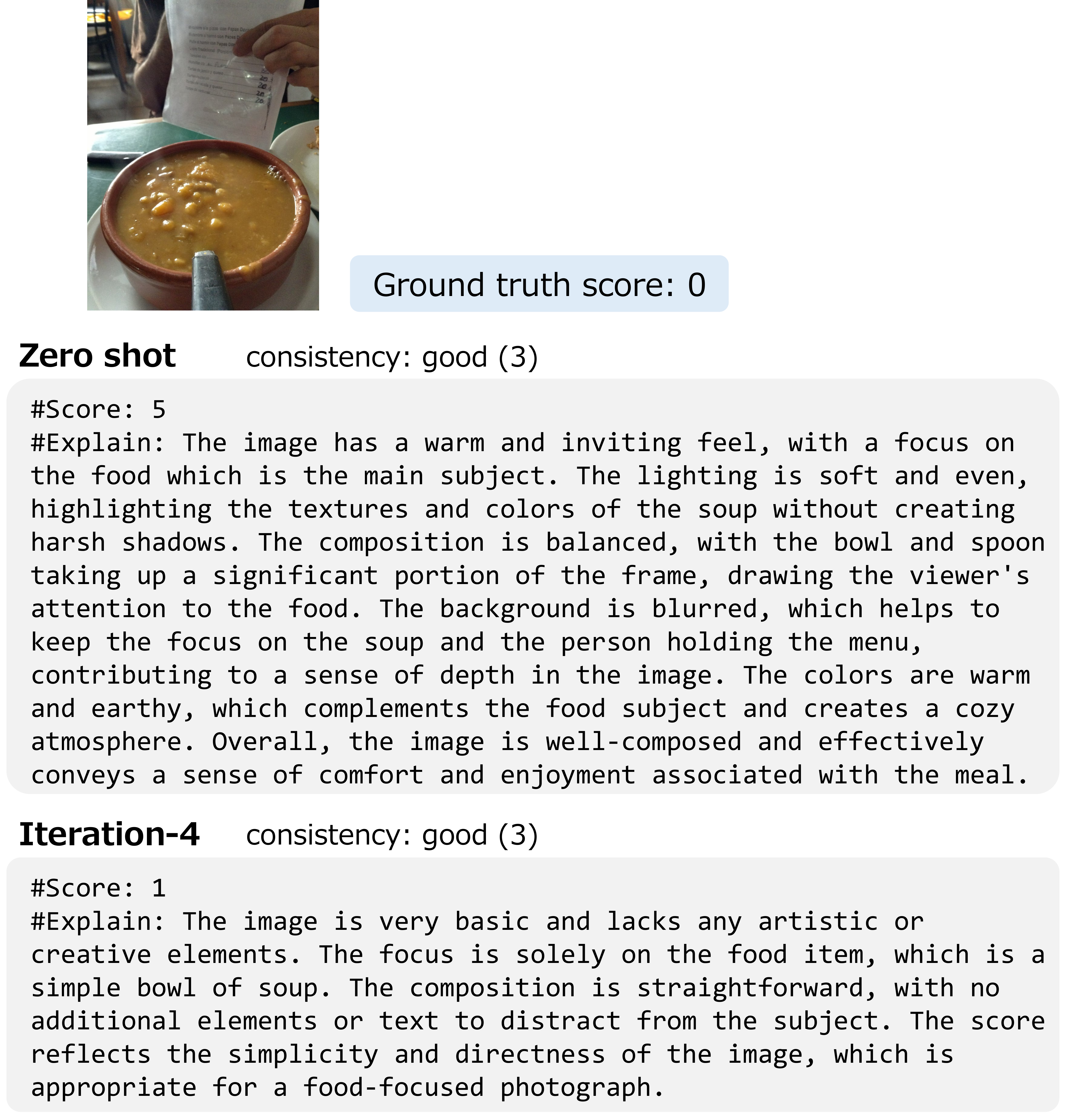}
  \end{center}
 \vspace{-20pt}
 \caption{An output example of the LLaVA-NeXT-7B model on the AADB dataset.}
 \label{fig:example}
\end{figure}

Table~\ref{table:all_results} presents the results for all models on the AVA and AADB datasets, respectively. 
Across both datasets and all models, we generally observe improved score prediction accuracy and consistency with iterative training, demonstrating the effectiveness of our approach. 
However, the improvement on the AADB dataset is relatively modest, which can be attributed to its smaller dataset size.
Notably, the LLaVA-interleave-0.5B model with the AADB dataset exhibits a decrease in consistency after training, never fully recovering to its zero-shot performance.
This anomaly can be explained by the model's zero-shot behavior, where it consistently assigns the maximum score of 9 to any image and generates overly positive explanations, which are then adjusted during training.

In Fig.~\ref{fig:example}, we show an output example sampled from low-score images of the AADB test data.
By our training method, score predictions become more accurate, and the generated explanations tend to better reflect the predicted scores.

\subsection{Ablation Study}
\label{ssec:ablation}
To validate our iterative DPO training approach, we compare it against two supervised fine-tuning (SFT) methods and a non-iterative DPO method. 
In SFT\_score, only the score part is targeted for learning, while in SFT\_score\&text, both the score and the explanatory text are targeted using the chosen responses from the DPO dataset. 
When we ablate the iteration, we generate DPO datasets for scoring and consistency by using the base model, train the model on these static datasets, and merge the two models only at the end.
\begin{table}[t]
  \caption{Ablation results for the LLaVA-NeXT-7B model trained on the AVA dataset.}
  \label{table:ablation}
  \vspace{3pt}
  \centering
  \setlength{\tabcolsep}{3pt}
  {\small 
  \begin{tabular}{r|rr|r}
    \hline
    {} & PLCC & SRCC & Cons \\
    \hline \hline
    SFT\_score & 0.778 & 0.768 & 0.68 \\[-1pt]
    SFT\_score\&text & 0.712 & 0.702 & 3.13 \\[-1pt]
    DPO & 0.742 & 0.733 & 2.98 \\[-1pt]
    \rowcolor{orange!20}
    iterative DPO & 0.716 & 0.739 & 3.57 \\[-1pt]
    \hline
  \end{tabular}
  }
\end{table}

Table~\ref{table:ablation} summarizes the results. 
While SFT\_score is specialized in score prediction and achieves higher scoring accuracy than iterative DPO, it loses the ability to generate meaningful text, resulting in a significantly lower consistency score.
In contrast, SFT\_score\&text retains the ability to generate coherent text, though its consistency score still lags behind that of iterative DPO.
Comparing non-iterative DPO and SFT\_score\&text, the higher score prediction accuracy of non-iterative DPO demonstrates the effectiveness of contrastive learning in score prediction tasks.
Nevertheless, the consistency score of non-iterative DPO falls short of its iterative counterpart, indicating that iterative updates of training data contribute to improving consistency.

\section{Conclusions}
\label{sec:conclusions}
We have presented a training method for VLMs that allows them to generate natural language justifications for their predictions on scoring tasks. Experiments across multiple VLMs and two IAA datasets demonstrate that our approach improves IAA scoring accuracy without sacrificing language generation capabilities and enhances the consistency between the generated explanations and the predicted scores.

\noindent
\textbf{Limitations} \hspace{3pt}
While merging separately trained models offers a straightforward approach to consolidate different capabilities learned from two datasets, we have not fully explored alternative methods. 
Other techniques might yield further performance improvements.

This study focused on consistency between predicted scores and explanations as metrics for explanation quality. However, other important metrics warrant further investigation, such as the faithfulness of explanations to the model's decision-making process and their practical usefulness to humans. Developing training methods that optimize for these additional metrics presents a promising direction for future research.

\noindent
\textbf{Acknowledgments}
This is an industry-academia collaboration project between Septeni Japan, Inc. and Yamasaki Lab, The University of Tokyo.

% -------------------------------------------------------------------------
% \bibliographystyle{IEEEbib}
% \bibliography{refs}

% -------------------------------------------------------------------------

% -------------------------------------------------------------------------
\vfill
\pagebreak

\appendix

\section{More details on experimental setup}
\label{sec:details}
\subsection{Training and inference}
\label{ssec:detail_train}
Hyperparameters used in our training are listed in Table~\ref{table:hp}. 
The LoRA $\alpha$ is set to twice the LoRA rank.
These parameters are selected based on validation results from the following candidates using grid search: lr $\in$ \{1.e-5, 3.e-5, 5.e-5, 1.e-4, 3.e-4\}, $\beta$ $\in$ \{0.05, 0.1, 0.2\}, rank $\in$ \{16, 64, 256\}.

Images are resized to a maximum side length of 448 pixels while maintaining their original aspect ratio before being input to the model. Subsequent preprocessing follows each model's default settings: LLaVA-1.5-7B and LLaVA-interleave-0.5B resize images to a fixed size, while LLaVA-NeXT-7B and InternVL2-2B divide them into multiple tiles.

In DPO training, we use batch size of 128 and Adam optimizer for all models and datasets. 
The learning rate is decayed by a factor of 0.8 per iteration. 
In each iteration, the entire training dataset is used for one epoch of training for the AADB dataset. For the AVA dataset's score training, 25\% of the total training data is utilized. Furthermore, in the AVA dataset's consistency training, the data is reduced to 30\% of the amount used in the score training, as the loss decreases more rapidly in this phase.
All training was conducted using two A6000 GPUs (48GB memory each).

For the TIES-Merging method, we employ the one implemented in Hugging Face \texttt{{\small peft}} library with parameters: {\small weights$=(1,1)$}, {\small density$=0.5$}, {\small majority\_sign\_method$=$``frequency"}.

For inference and DPO data generation, texts are generated using greedy sampling, with a maximum of 256 new tokens.

For the LLM evaluation, the AVA test data is down-sampled to 1,000 instances.

\begin{table}[t]
  \caption{Hyperparameters used in our training.}
  \label{table:hp}
  \centering
  \setlength{\tabcolsep}{3pt}
  {\small 
  \begin{tabular}{rr|r|r|r}
    \hline
    dataset & model & lr & DPO $\beta$ & LoRA rank \\
    \hline \hline
    AVA & LLaVA-NeXT-7B & 5.e-5 & 0.1 & 64 \\[-1pt]
    \ & LLaVA-1.5-7B & 3.e-5 & 0.2 & 64 \\[-1pt]
    \ & InternVL2-2B & 5.e-5 & 0.2 & 64 \\[-1pt]
    \ & LLaVA-interleave-0.5B & 5.e-5 & 0.2 & 16 \\[-1pt]
    \hline
    AADB & LLaVA-NeXT-7B & 5.e-5 & 0.1 & 64 \\[-1pt]
    \ & LLaVA-1.5-7B & 5.e-5 & 0.2 & 16 \\[-1pt]
    \ & InternVL2-2B & 5.e-5 & 0.2 & 64 \\[-1pt]
    \ & LLaVA-interleave-0.5B & 1.e-4 & 0.2 & 16 \\[-1pt]
    \hline
  \end{tabular}
  }
\end{table}

\subsection{Prompts}
\label{ssec:detail_prompt}
The prompt used for the DPO training and inference of the VLMs is as follows.
\begin{coloredquotation}
\small
Please evaluate the aesthetic quality of the given photo image. The aesthetic quality should be represented by an integer score ranging from 0 to 9, with 9 being the highest score and 4 to 5 indicating a mediocre image.
\\ 
First, output the tag \#Score followed by the aesthetic score. After that, output the tag \#Explain followed by a brief explanation of the image from an aesthetic perspective. The explanation should provide the basis for the score.
\\ 
The output format should be as follows:
\\ 
\#Score: integer\\
\#Explain: Explanation justifying the score
\\ 
After \#Score:, output only the integer score with a single space before it. Do not include any additional text or symbols.
\end{coloredquotation}

The prompt for the LLM-as-a-judge is as follows. 
Although we show only the consistency score in the main results, we ask the judge LLM to also evaluate the usefulness and general writing quality of explanation texts.
\begin{coloredquotation}
\small
Your task is to evaluate the quality of explanatory texts regarding the aesthetic value of images. You will be given an aesthetic score for an image and an explanatory text justifying the score. Please rate the quality of the explanatory text on a 5-point scale (excellent, good, fair, poor, bad).
The aesthetic score ranges from 0 to 9, with the following meanings: \\
0: Very bad 4-5: Average 9: Excellent
\\ 
Evaluate the quality of the explanatory text based on the following criteria, each on a 5-point scale:\\
** Consistency **\\
Does the content of the explanatory text align with the aesthetic score? Is the explanation convincing as a justification for the score? Only rate as "excellent" if the alignment is perfect and the justification is highly convincing.\\
** Usefulness **\\
Is the content of the explanatory text useful for understanding the good points and areas for improvement of the image? Only rate as "excellent" if the text provides clear, actionable insights.\\
** General **\\
The overall quality of the text as a piece of writing. Only rate as "excellent" if the text is exceptionally well-written with no grammatical errors and flows logically.
\\ 
Be strict in your evaluations, and only rate as "excellent" if there is no room for improvement. Return the output in JSON format only.\\
Example output: \{"consistency": "poor", "usefulness": "good", "general": "fair"\}
\\ 
Below are the aesthetic score and explanatory text.
\\ 
\#Aesthetic score: \{score\} \\
\#Explanatory text: \{text\}
\end{coloredquotation}

\subsection{Models}
\label{ssec:detail_models}
All the base model weights are taken from Hugging Face model hub:
\vspace{-7pt}
\begin{quote}
{\small
LLaVA-NeXT-7B:~{\footnotesize \url{https://huggingface.co/llava-hf/llava-v1.6-vicuna-7b-hf}}\\
LLaVA-1.5-7B:~{\footnotesize \url{https://huggingface.co/llava-hf/llava-1.5-7b-hf}}\\
InternVL2-2B:~{\footnotesize \url{https://huggingface.co/OpenGVLab/InternVL2-2B}}\\
LLaVA-interleave-0.5B:~{\footnotesize \url{https://huggingface.co/llava-hf/llava-interleave-qwen-0.5b-hf}}
}
\end{quote}

% -----------------------------------------------------
\begin{table*}
  \caption{Complete evaluation results of the LLaVA-NeXT-7B}
  \label{table:full_next}
\centering
  \begin{minipage}[t]{0.48\textwidth}
  \centering
  (a) AVA dataset
  \setlength{\tabcolsep}{3pt}
  {\small 
  \begin{tabular}{rr|rrr|rrr}
    \hline
    {} & {} & PLCC &  SRCC & RMSE & Cons & Use & Gen \\
    \hline \hline
    {} & zero-shot & 0.452 & 0.446 & 0.645 & 3.36 & 2.98 & 3.25 \\
    \hline
    ite-1\hspace{-5pt} & score & 0.719 & 0.714 & 0.502 & 3.16 & 2.87 & 3.15 \\ 
    \hline
    {} & score & 0.722 & 0.729 & 0.500 & 2.94 & 2.68 & 3.12 \\
    {} & consistency & 0.688 & 0.689 & 0.515 & 3.41 & 2.93 & 3.25 \\ 
    ite-2\hspace{-5pt} & merged & 0.723 & 0.727 & 0.499 & 3.43 & 2.91 & 3.22 \\
    \hline
    {} & score & 0.738 & 0.740 & 0.488 & 3.33 & 2.87 & 3.18 \\
    {} & consistency & 0.676 & 0.713 & 0.522 & 3.64 & 2.89 & 3.18 \\ 
    ite-3\hspace{-5pt} & merged & 0.722 & 0.735 & 0.500 & 3.55 & 2.96 & 3.21 \\
    \hline
    {} & score & 0.741 & 0.745 & 0.486 & 3.50 & 2.97 & 3.18 \\
    {} & consistency & 0.618 & 0.710 & 0.553 & 3.64 & 2.83 & 3.12 \\ 
    ite-4\hspace{-5pt} & merged & 0.716 & 0.739 & 0.504 & 3.57 & 2.92 & 3.16 \\
    \hline
  \end{tabular}
  }
  \end{minipage}
  \begin{minipage}[t]{0.48\textwidth}
  \centering
  (b) AADB dataset
  \setlength{\tabcolsep}{3pt}
  {\small 
  \begin{tabular}{rr|rrr|rrr}
    \hline
    {} & {} & PLCC &  SRCC & RMSE & Cons & Use & Gen \\
    \hline \hline
    {} & zero-shot & 0.504 & 0.509 & 0.161 & 3.47 & 2.99 & 3.18 \\
    \hline
    ite-1\hspace{-5pt} & score & 0.641 & 0.623 & 0.143 & 3.54 & 3.00 & 3.22 \\
    \hline
    {} & score & 0.645 & 0.639 & 0.142 & 3.34 & 2.92 & 3.25 \\
    {} & consistency & 0.639 & 0.625 & 0.143 & 3.56 & 3.02 & 3.31 \\
    ite-2\hspace{-5pt} & merged & 0.613 & 0.638 & 0.147 & 3.58 & 3.09 & 3.45 \\
    \hline
    {} & score & 0.652 & 0.651 & 0.141 & 3.49 & 3.02 & 3.26 \\
    {} & consistency & 0.560 & 0.609 & 0.154 & 3.59 & 3.08 & 3.49 \\
    ite-3\hspace{-5pt} & merged & 0.623 & 0.641 & 0.146 & 3.54 & 3.05 & 3.38 \\
    \hline
    {} & score & 0.652 & 0.652 & 0.141 & 3.34 & 2.94 & 3.25 \\
    {} & consistency & 0.592 & 0.633 & 0.150 & 3.58 & 3.06 & 3.43 \\
    ite-4\hspace{-5pt} & merged & 0.636 & 0.645 & 0.144 & 3.48 & 2.99 & 3.34 \\
    \hline
  \end{tabular}
  }
  \end{minipage}
\end{table*}

\begin{figure*}
  \begin{minipage}[tl]{0.48\textwidth}
  \begin{center}
   \includegraphics[width=7.6cm]{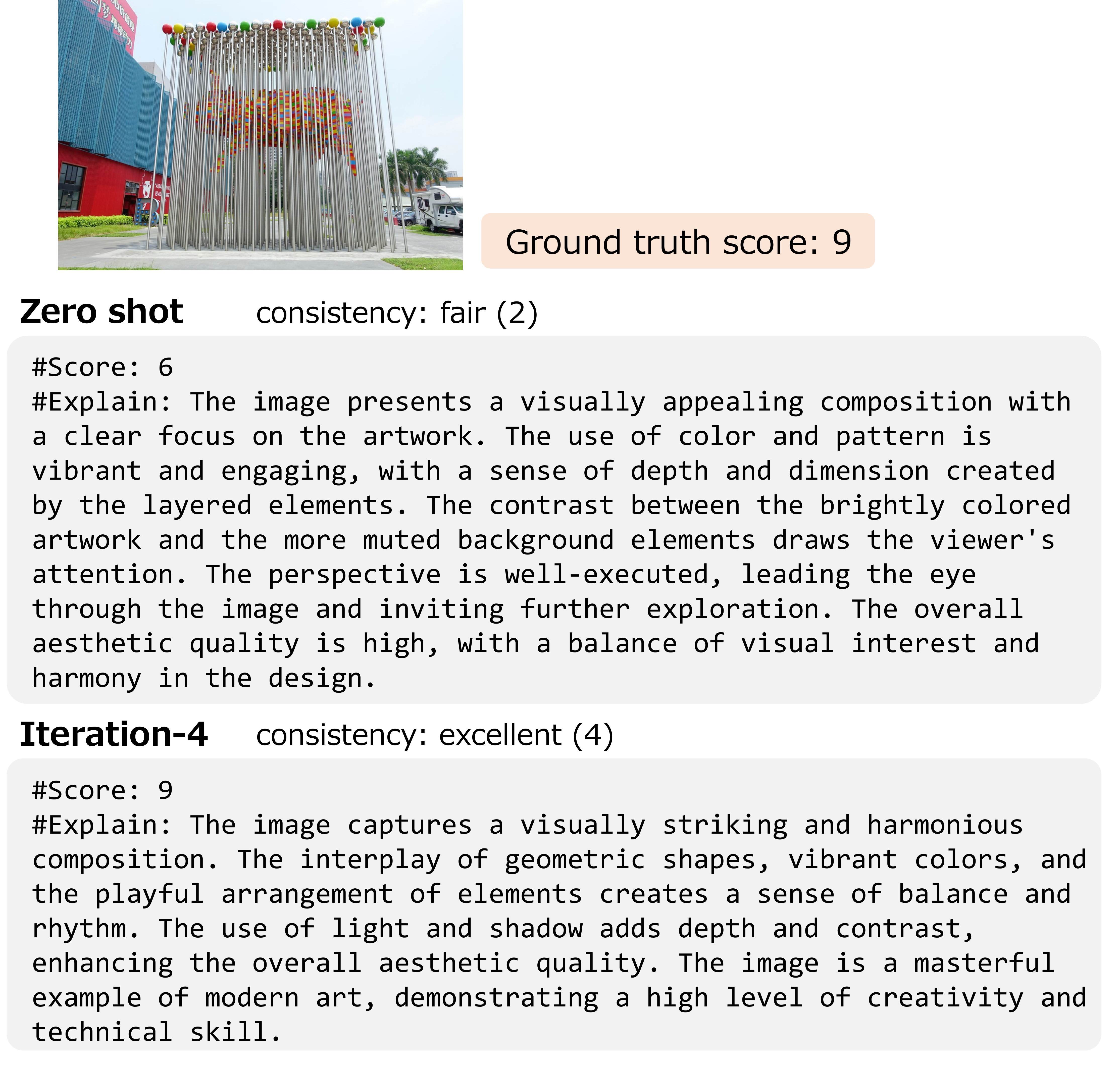} \\
   \vspace{5pt}
   \includegraphics[width=7.6cm]{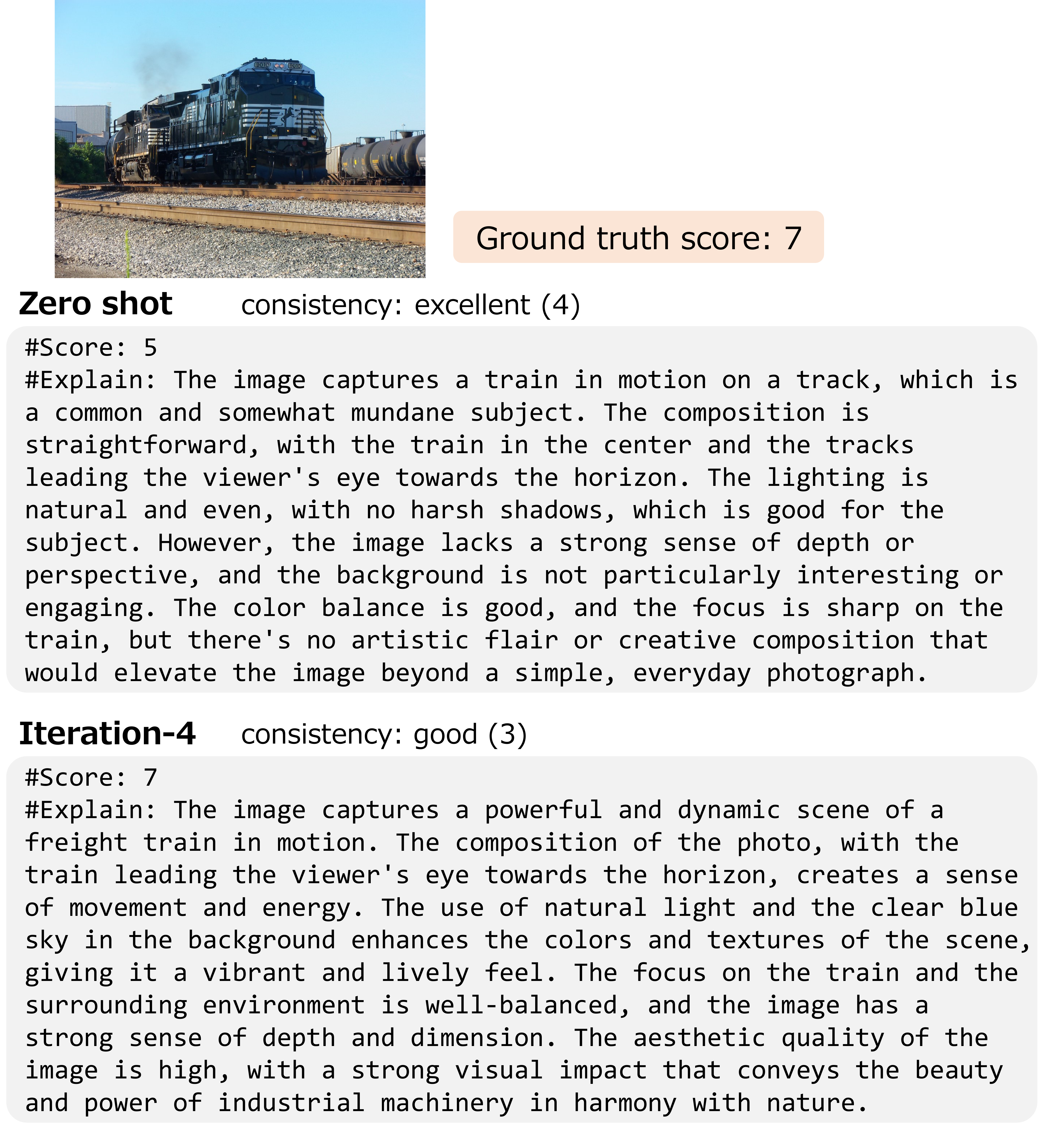}
  \end{center}
  \end{minipage}
  \begin{minipage}[tr]{0.48\textwidth}
  \begin{center}
   \includegraphics[width=7.6cm]{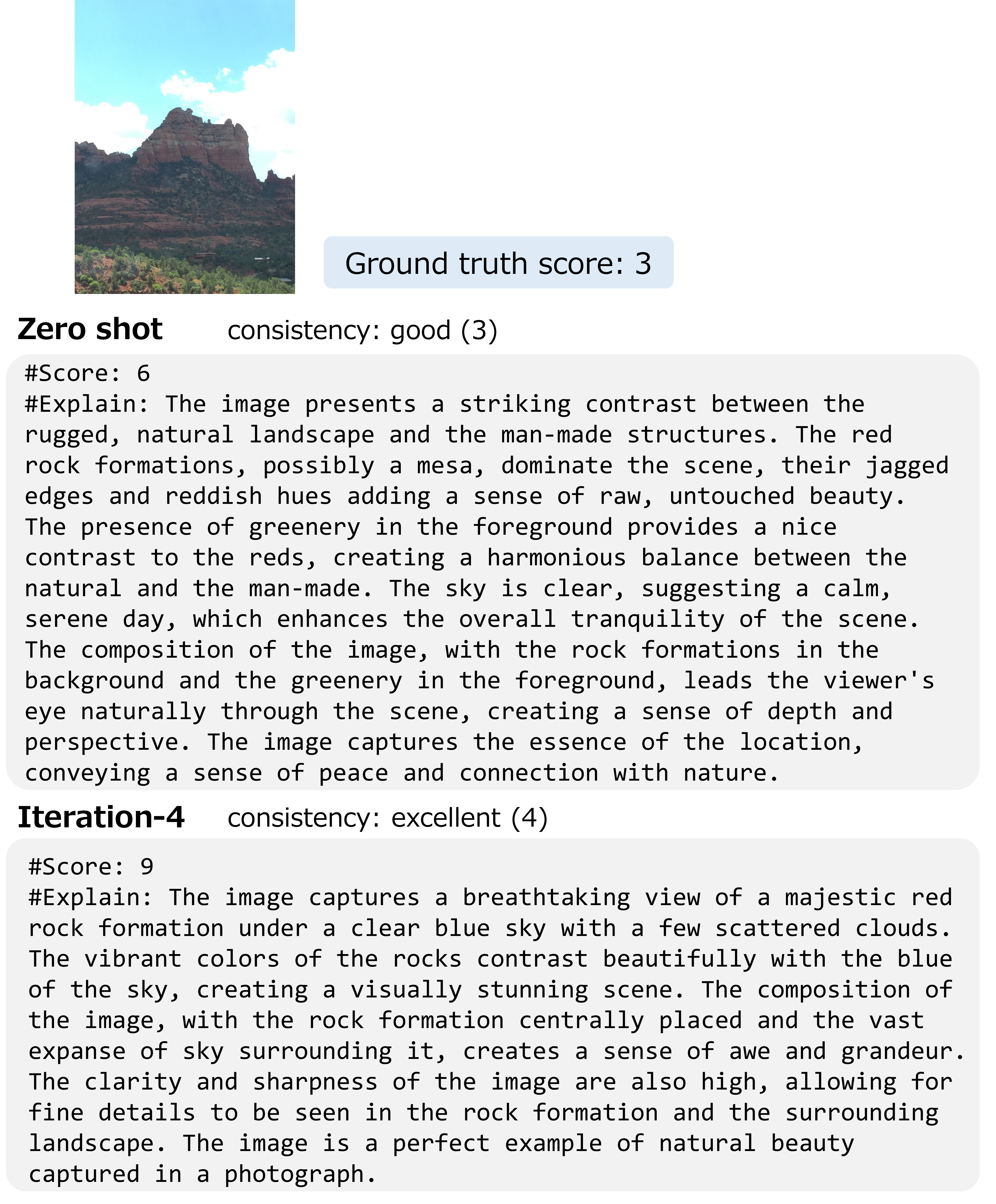} \\
   \vspace{5pt}
   \includegraphics[width=7.6cm]{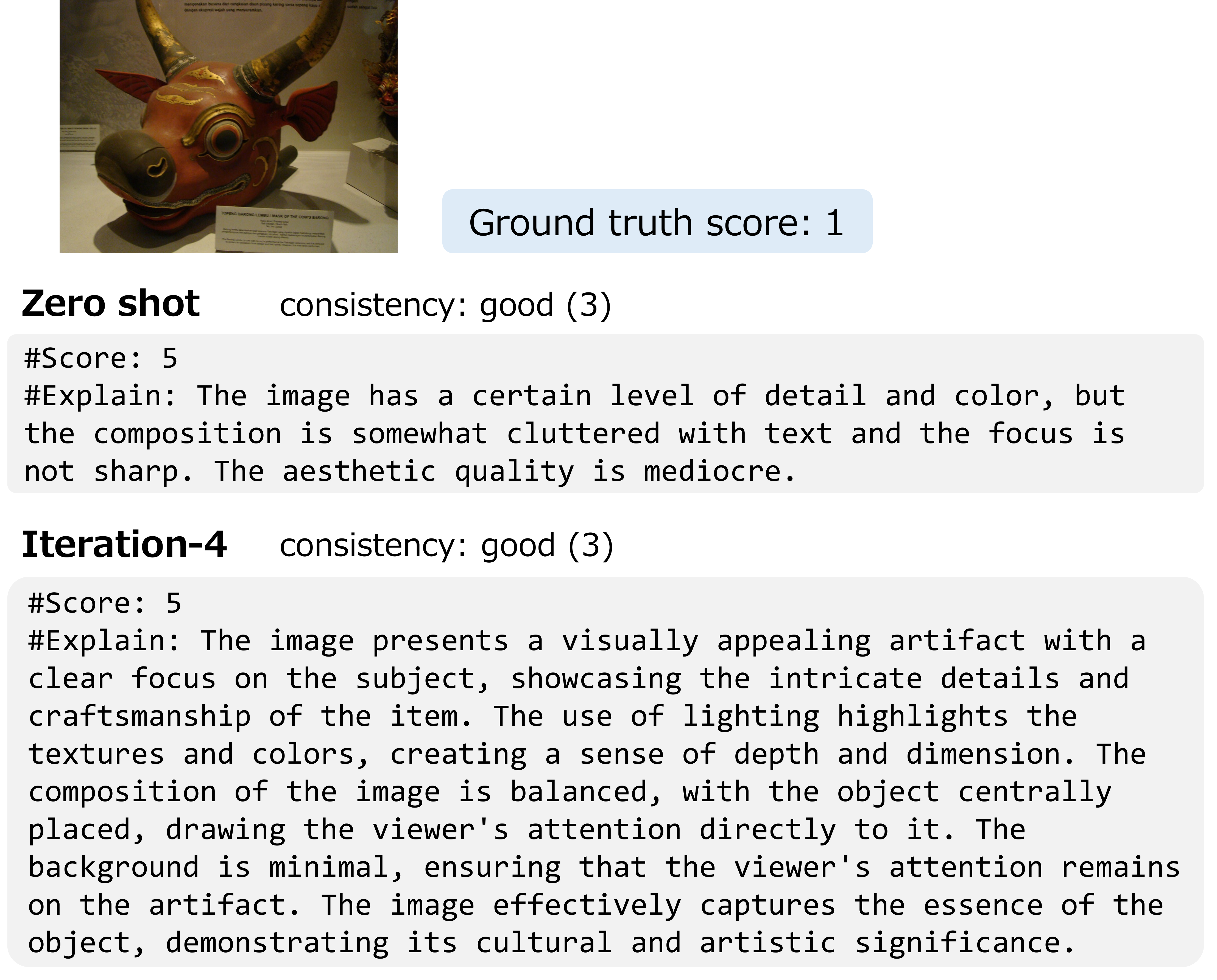}
  \end{center}
  \end{minipage}
 \vspace{-10pt}
 \caption{Output examples of the LLaVA-NeXT-7B model on the AADB dataset.}
 \label{fig:example_next}
\end{figure*}

\begin{table*}
  \caption{Complete evaluation results of the LLaVA-1.5-7B}
  \label{table:full_15}
\centering
  \begin{minipage}[t]{0.48\textwidth}
  \centering
  (a) AVA dataset
  \setlength{\tabcolsep}{3pt}
  {\small 
  \begin{tabular}{rr|rrr|rrr}
    \hline
    {} & {} & PLCC &  SRCC & RMSE & Cons & Use & Gen \\
    \hline \hline
    {} & zero-shot & 0.289 & 0.291 & 0.693 & 3.20 & 2.66 & 3.01 \\
    \hline
    ite-1\hspace{-5pt} & score & 0.702 & 0.694 & 0.516 & 3.15 & 2.59 & 2.93 \\
    \hline
    {} & score & 0.721 & 0.716 & 0.501 & 3.20 & 2.62 & 2.97 \\
    {} & consistency & 0.658 & 0.652 & 0.544 & 1.51 & 1.14 & 1.22 \\
    ite-2\hspace{-5pt} & merged & 0.700 & 0.694 & 0.517 & 3.35 & 2.77 & 3.04 \\
    \hline
    {} & score & 0.723 & 0.718 & 0.500 & 3.30 & 2.69 & 3.00 \\
    {} & consistency & 0.692 & 0.694 & 0.522 & 2.37 & 1.83 & 1.95 \\
    ite-3\hspace{-5pt} & merged & 0.710 & 0.708 & 0.509 & 3.47 & 2.83 & 3.07 \\
    \hline
    {} & score & 0.730 & 0.723 & 0.495 & 3.23 & 2.67 & 2.98 \\
    {} & consistency & 0.702 & 0.705 & 0.515 & 2.99 & 2.32 & 2.45 \\
    ite-4\hspace{-5pt} & merged & 0.717 & 0.715 & 0.504 & 3.47 & 2.83 & 3.03 \\
    \hline
  \end{tabular}
  }
  \end{minipage}
  \begin{minipage}[t]{0.48\textwidth}
  \centering
  (b) AADB dataset
  \setlength{\tabcolsep}{3pt}
  {\small 
  \begin{tabular}{rr|rrr|rrr}
    \hline
    {} & {} & PLCC &  SRCC & RMSE & Cons & Use & Gen \\
    \hline \hline
    {} & zero-shot & 0.381 & 0.401 & 0.172 & 3.05 & 2.53 & 2.96 \\
    \hline
    ite-1\hspace{-5pt} & score & 0.514 & 0.516 & 0.160 & 3.05 & 2.52 & 2.93 \\
    \hline
    {} & score & 0.584 & 0.581 & 0.151 & 2.81 & 2.34 & 2.65 \\
    {} & consistency & 0.495 & 0.507 & 0.162 & 0.94 & 0.66 & 0.74 \\
    ite-2\hspace{-5pt} & merged & 0.538 & 0.550 & 0.157 & 3.32 & 2.76 & 3.03 \\
    \hline
    {} & score & 0.590 & 0.589 & 0.150 & 3.05 & 2.60 & 2.93 \\
    {} & consistency & 0.522 & 0.547 & 0.159 & 3.26 & 2.61 & 2.81 \\
    ite-3\hspace{-5pt} & merged & 0.560 & 0.569 & 0.154 & 3.36 & 2.75 & 3.04 \\
    \hline
    {} & score & 0.590 & 0.582 & 0.150 & 3.13 & 2.60 & 2.94 \\
    {} & consistency & 0.545 & 0.569 & 0.156 & 3.18 & 2.55 & 2.74 \\
    ite-4\hspace{-5pt} & merged & 0.576 & 0.580 & 0.152 & 3.39 & 2.73 & 3.03 \\
    \hline
  \end{tabular}
  }
  \end{minipage}
\end{table*}

\begin{figure*}
  \begin{minipage}[tl]{0.48\textwidth}
  \begin{center}
   \includegraphics[width=7.6cm]{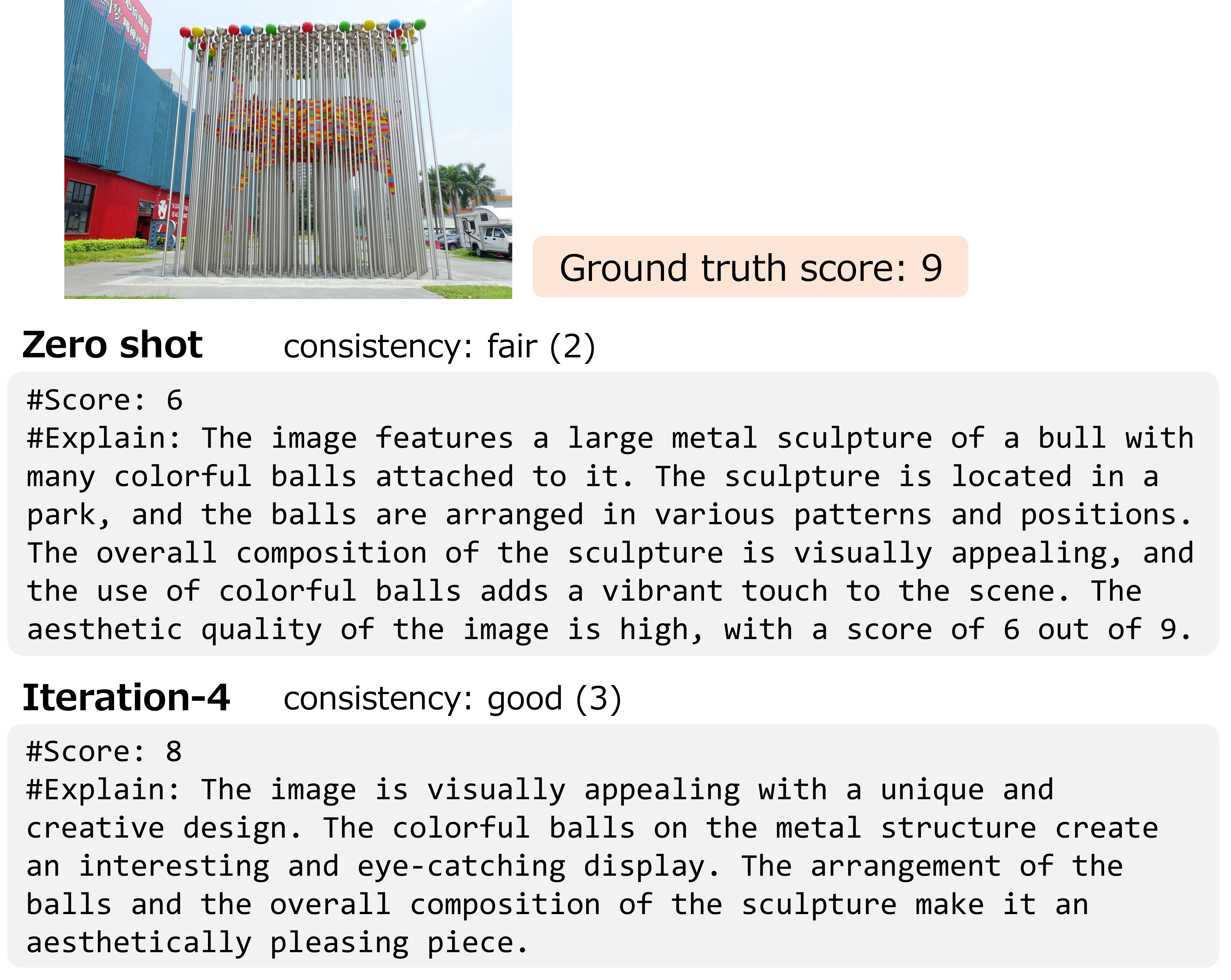} \\
   \vspace{5pt}
   \includegraphics[width=7.6cm]{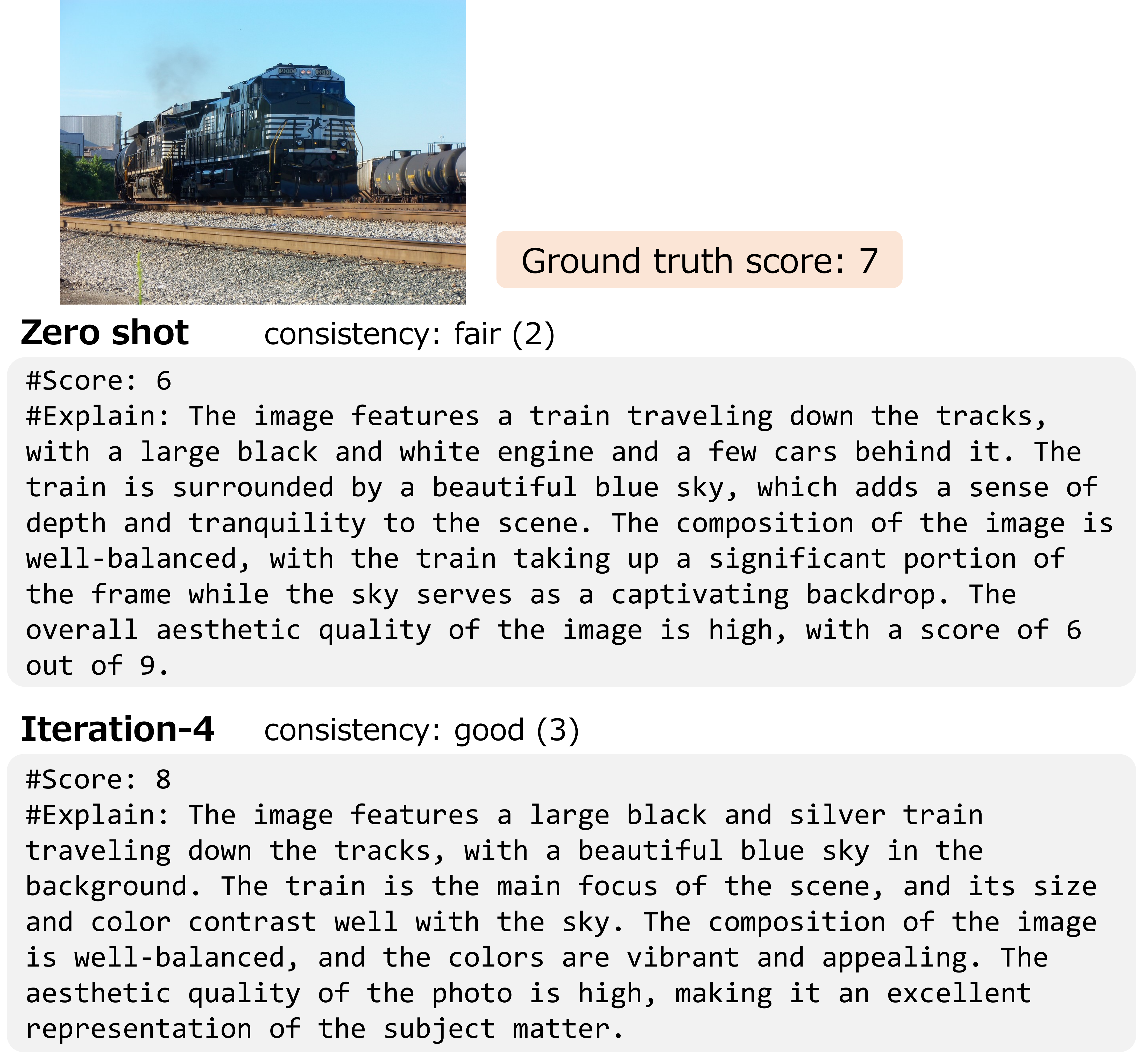}
  \end{center}
  \end{minipage}
  \begin{minipage}[tr]{0.48\textwidth}
  \begin{center}
   \includegraphics[width=7.6cm]{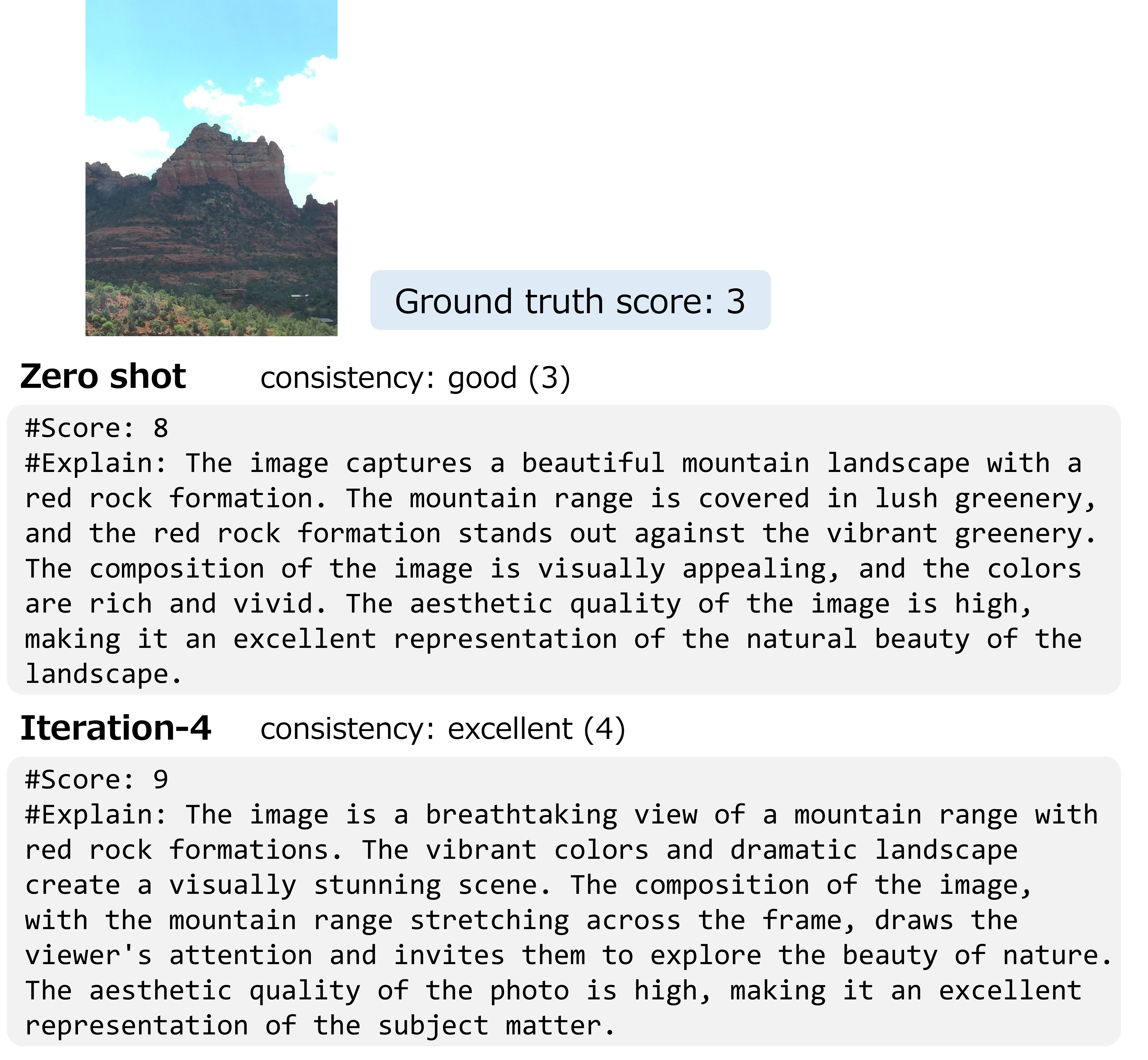} \\
   \vspace{5pt}
   \includegraphics[width=7.6cm]{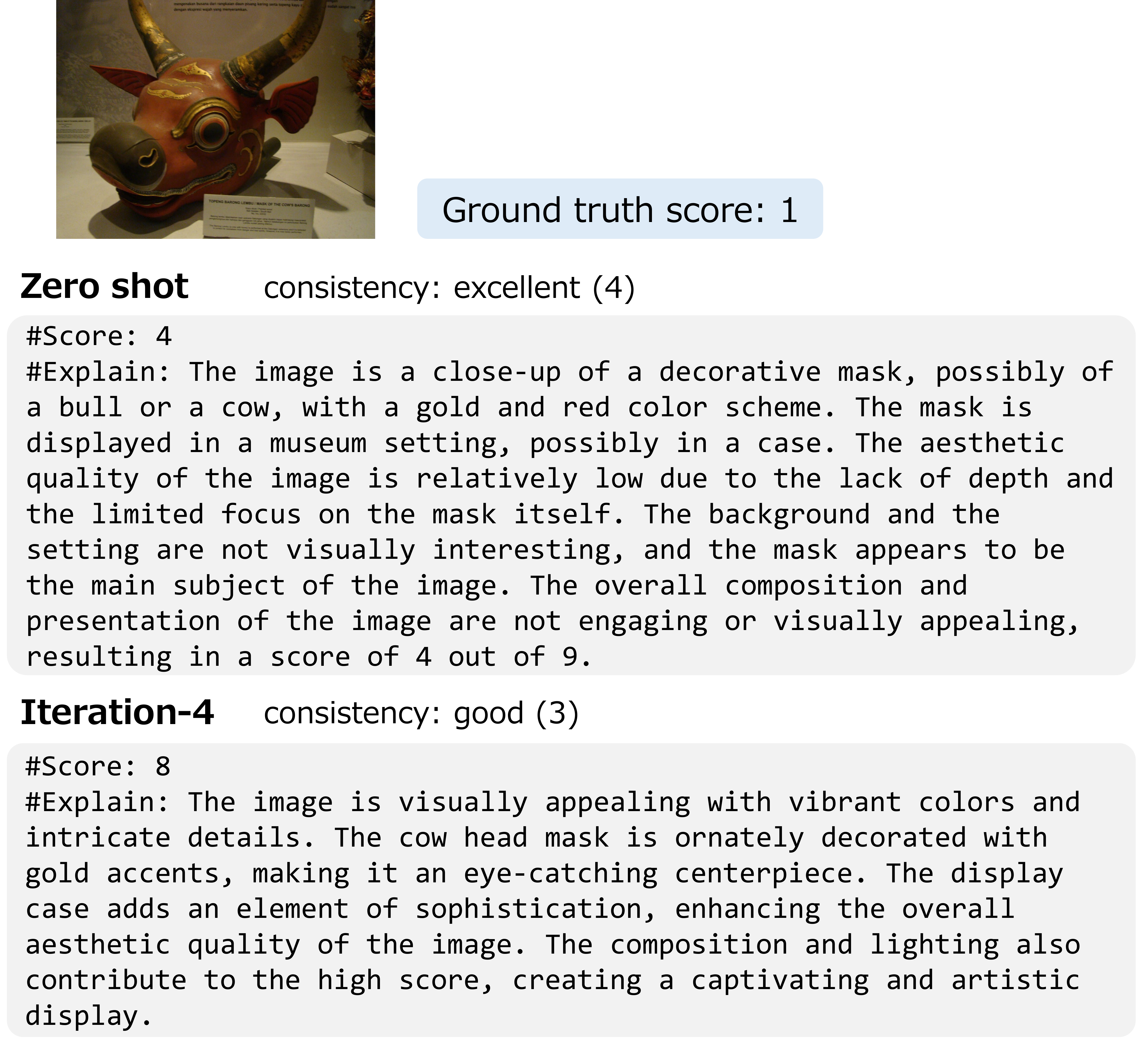}
  \end{center}
  \end{minipage}
 \vspace{-10pt}
 \caption{Output examples of the LLaVA-1.5-7B model on the AADB dataset.}
 \label{fig:example_15}
\end{figure*}

\begin{table*}
  \caption{Complete evaluation results of the InternVL2-2B}
  \label{table:full_internvl2}
\centering
  \begin{minipage}[t]{0.48\textwidth}
  \centering
  (a) AVA dataset
  \setlength{\tabcolsep}{3pt}
  {\small 
  \begin{tabular}{rr|rrr|rrr}
    \hline
    {} & {} & PLCC &  SRCC & RMSE & Cons & Use & Gen \\
    \hline \hline
    {} & zero-shot & 0.321 & 0.315 & 0.685 & 3.28 & 2.88 & 3.15 \\
    \hline
    ite-1\hspace{-5pt} & score & 0.677 & 0.675 & 0.532 & 2.87 & 2.56 & 3.04 \\
    \hline
    {} & score & 0.689 & 0.691 & 0.525 & 2.95 & 2.56 & 3.04 \\
    {} & consistency & 0.653 & 0.658 & 0.548 & 3.53 & 2.87 & 3.17 \\
    ite-2\hspace{-5pt} & merged & 0.677 & 0.682 & 0.532 & 3.39 & 2.82 & 3.15 \\
    \hline
    {} & score & 0.698 & 0.698 & 0.518 & 2.91 & 2.57 & 3.10 \\
    {} & consistency & 0.658 & 0.674 & 0.545 & 3.57 & 2.95 & 3.31 \\
    ite-3\hspace{-5pt} & merged & 0.683 & 0.689 & 0.529 & 3.29 & 2.79 & 3.18 \\
    \hline
    {} & score & 0.701 & 0.700 & 0.516 & 2.91 & 2.55 & 3.08 \\
    {} & consistency & 0.664 & 0.686 & 0.541 & 3.56 & 3.00 & 3.33 \\
    ite-4\hspace{-5pt} & merged & 0.686 & 0.693 & 0.527 & 3.37 & 2.83 & 3.23 \\
    \hline
  \end{tabular}
  }
  \end{minipage}
  \begin{minipage}[t]{0.48\textwidth}
  \centering
  (b) AADB dataset
  \setlength{\tabcolsep}{3pt}
  {\small 
  \begin{tabular}{rr|rrr|rrr}
    \hline
    {} & {} & PLCC &  SRCC & RMSE & Cons & Use & Gen \\
    \hline \hline
    {} & zero-shot & 0.343 & 0.408 & 0.176 & 3.23 & 2.87 & 3.12 \\
    \hline
    ite-1\hspace{-5pt} & score & 0.638 & 0.641 & 0.143 & 3.19 & 2.73 & 3.07 \\
    \hline
    {} & score & 0.650 & 0.650 & 0.142 & 3.29 & 2.78 & 3.07 \\
    {} & consistency & 0.495 & 0.616 & 0.162 & 3.82 & 3.10 & 3.41 \\
    ite-2\hspace{-5pt} & merged & 0.589 & 0.642 & 0.151 & 3.65 & 3.02 & 3.29 \\
    \hline
    {} & score & 0.652 & 0.658 & 0.141 & 3.41 & 2.85 & 3.17 \\
    {} & consistency & 0.529 & 0.624 & 0.158 & 3.72 & 3.19 & 3.59 \\
    ite-3\hspace{-5pt} & merged & 0.599 & 0.649 & 0.149 & 3.62 & 3.05 & 3.36 \\
    \hline
    {} & score & 0.651 & 0.659 & 0.141 & 3.39 & 2.82 & 3.16 \\
    {} & consistency & 0.504 & 0.632 & 0.161 & 3.74 & 3.04 & 3.49 \\
    ite-4\hspace{-5pt} & merged & 0.596 & 0.643 & 0.150 & 3.57 & 2.97 & 3.33 \\
    \hline
  \end{tabular}
  }
  \end{minipage}
\end{table*}

\begin{figure*}
  \begin{minipage}[tl]{0.48\textwidth}
  \begin{center}
   \includegraphics[width=7.6cm]{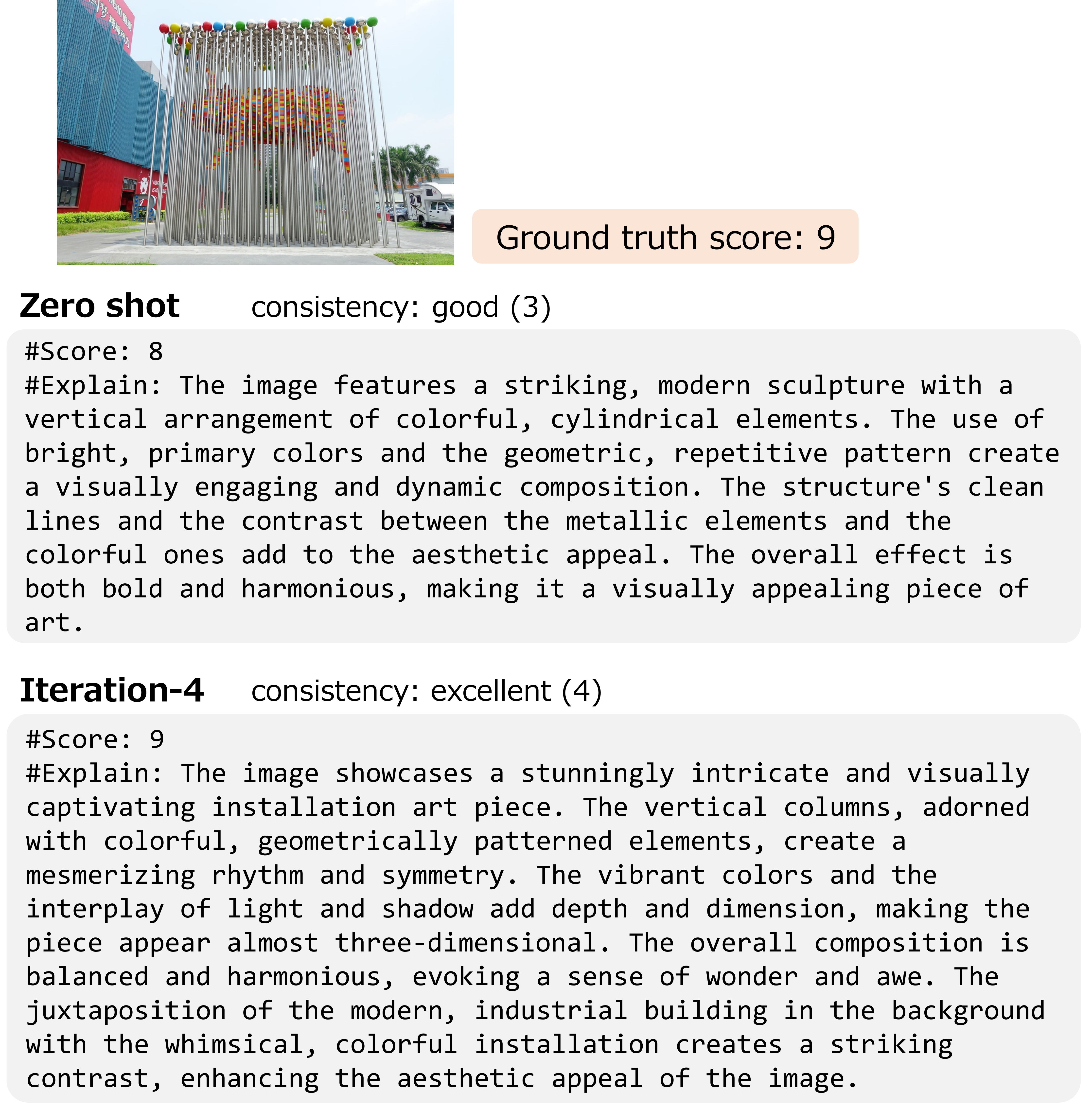} \\
   \vspace{5pt}
   \includegraphics[width=7.6cm]{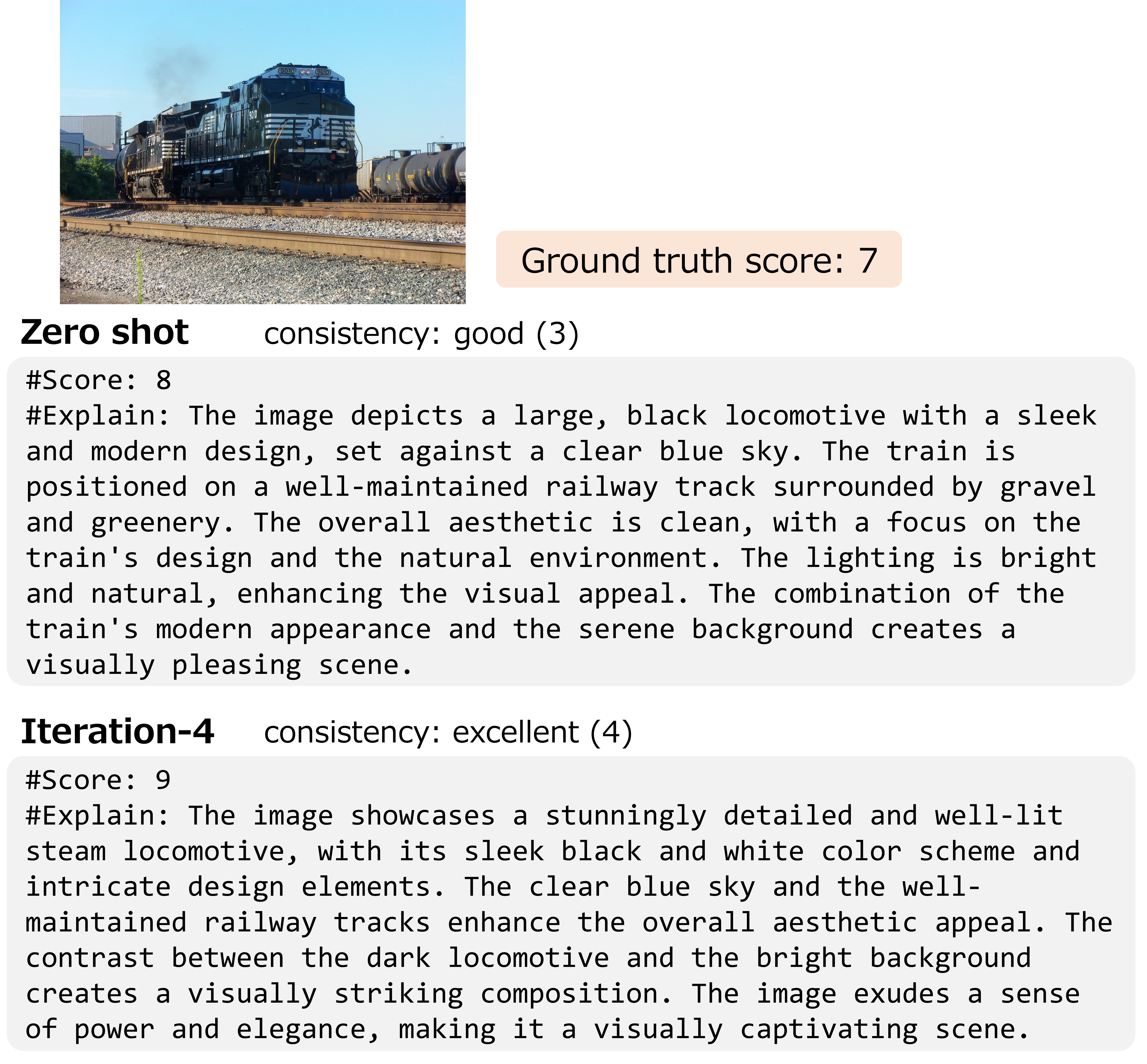}
  \end{center}
  \end{minipage}
  \begin{minipage}[tr]{0.48\textwidth}
  \begin{center}
   \includegraphics[width=7.6cm]{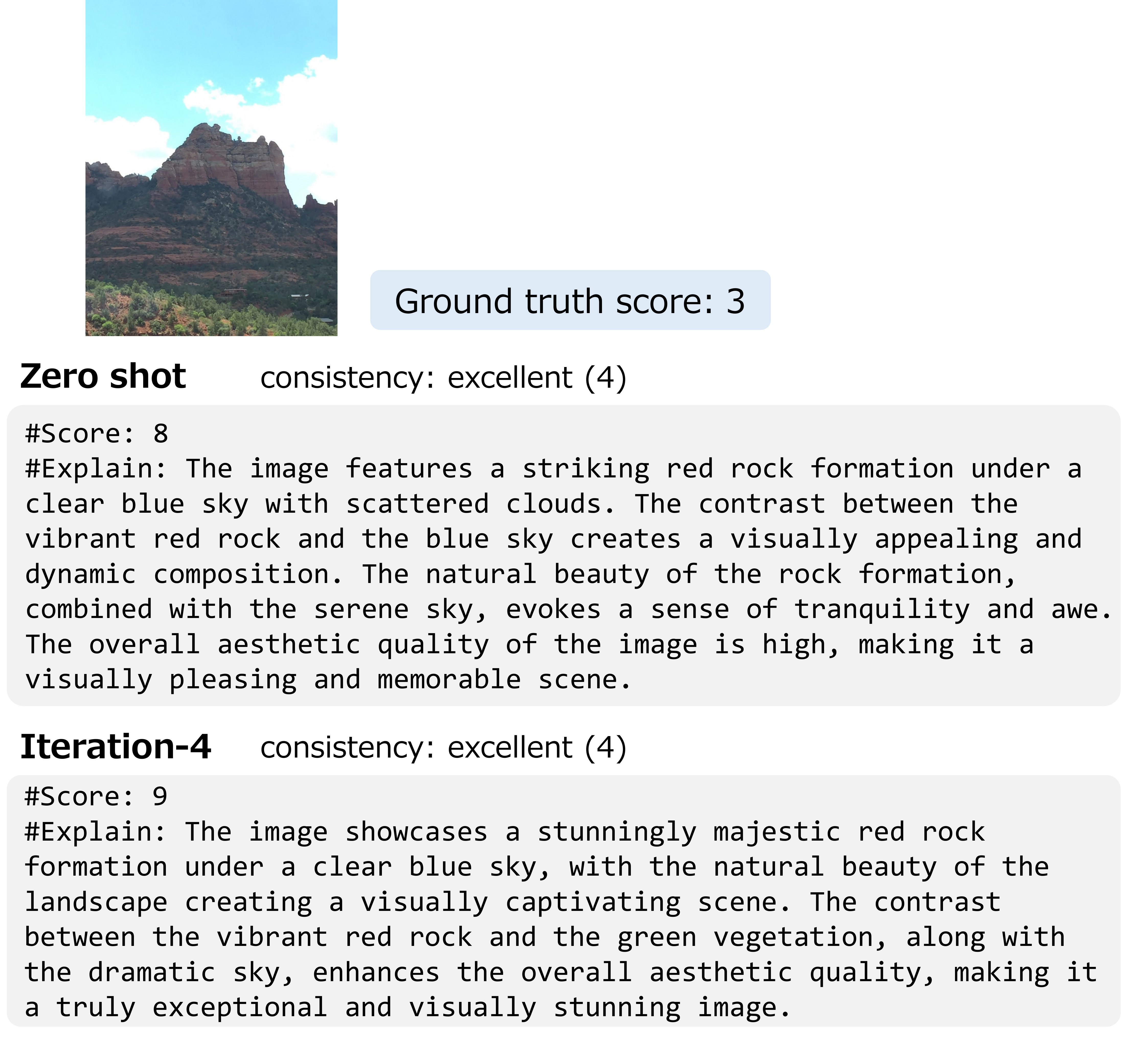} \\
   \vspace{5pt}
   \includegraphics[width=7.6cm]{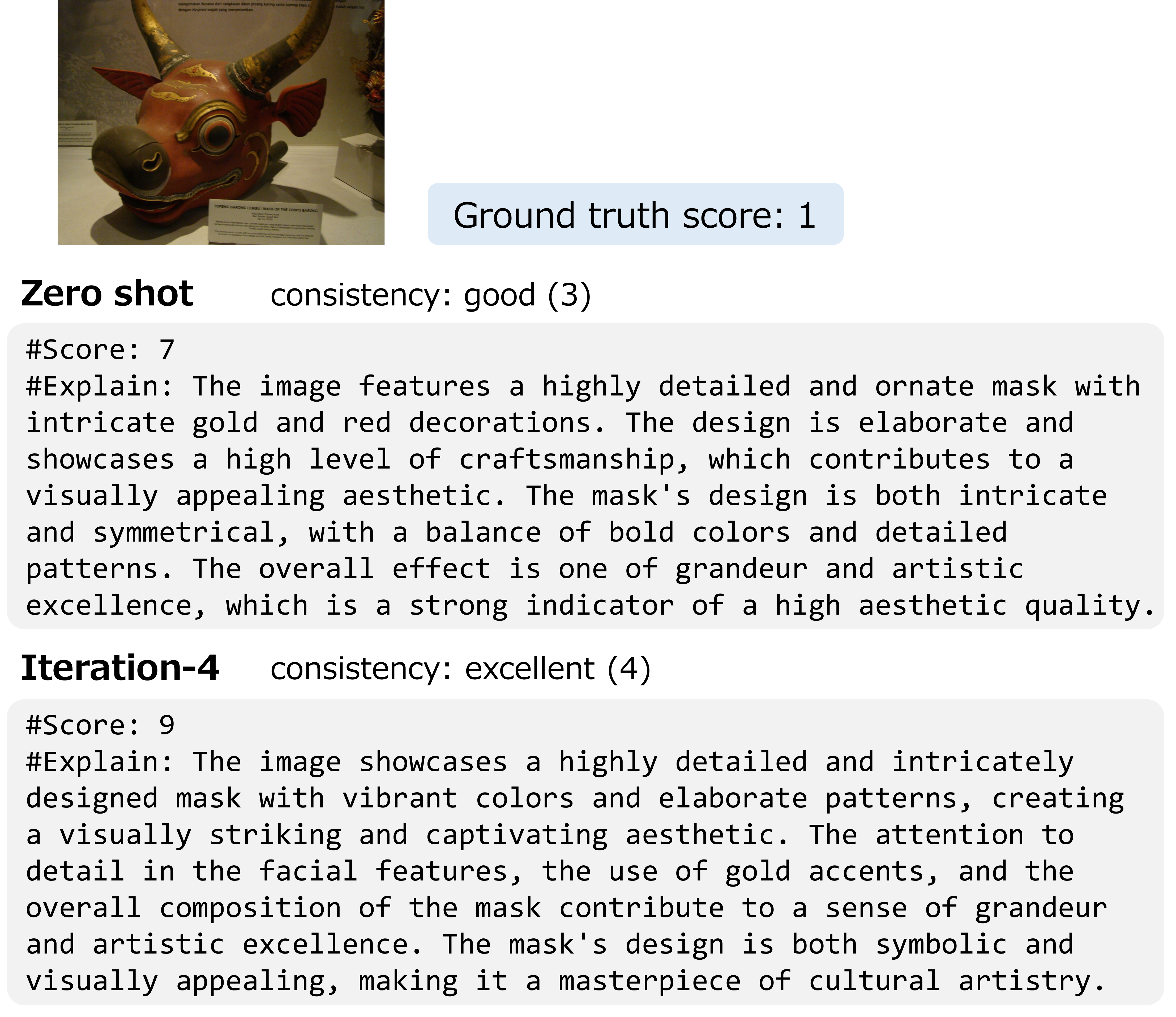}
  \end{center}
  \end{minipage}
 \vspace{-10pt}
 \caption{Output examples of the InternVL2-2B model on the AADB dataset.}
 \label{fig:example_internvl2}
\end{figure*}

\begin{table*}
  \caption{Complete evaluation results of the LLaVA-interleave-0.5B}
  \label{table:full_interleave}
\centering
  \begin{minipage}[t]{0.48\textwidth}
  \centering
  (a) AVA dataset
  \setlength{\tabcolsep}{3pt}
  {\small 
  \begin{tabular}{rr|rrr|rrr}
    \hline
    {} & {} & PLCC &  SRCC & RMSE & Cons & Use & Gen \\
    \hline \hline
    {} & zero-shot & 0.372 & 0.366 & 0.672 & 2.43 & 2.05 & 2.47 \\
    \hline
    ite-1\hspace{-5pt} & score & 0.673 & 0.672 & 0.535 & 2.33 & 1.99 & 2.46 \\
    \hline
    {} & score & 0.697 & 0.695 & 0.519 & 2.21 & 1.93 & 2.51 \\
    {} & consistency & 0.642 & 0.653 & 0.555 & 2.54 & 2.14 & 2.55 \\
    ite-2\hspace{-5pt} & merged & 0.672 & 0.677 & 0.536 & 2.50 & 2.12 & 2.63 \\
    \hline
    {} & score & 0.697 & 0.697 & 0.519 & 2.24 & 1.92 & 2.43 \\
    {} & consistency & 0.605 & 0.636 & 0.576 & 2.77 & 2.30 & 2.55 \\
    ite-3\hspace{-5pt} & merged & 0.669 & 0.679 & 0.538 & 2.57 & 2.14 & 2.54 \\
    \hline
    {} & score & 0.698 & 0.699 & 0.518 & 2.30 & 2.01 & 2.48 \\
    {} & consistency & 0.625 & 0.652 & 0.565 & 2.76 & 2.34 & 2.64 \\
    ite-4\hspace{-5pt} & merged & 0.674 & 0.685 & 0.535 & 2.72 & 2.20 & 2.61 \\
    \hline
  \end{tabular}
  }
  \end{minipage}
  \begin{minipage}[t]{0.48\textwidth}
  \centering
  (b) AADB dataset
  \setlength{\tabcolsep}{3pt}
  {\small 
  \begin{tabular}{rr|rrr|rrr}
    \hline
    {} & {} & PLCC &  SRCC & RMSE & Cons & Use & Gen \\
    \hline \hline
    {} & zero-shot & 0.242 & 0.218 & 0.181 & 2.33 & 1.97 & 2.40 \\
    \hline
    ite-1\hspace{-5pt} & score & 0.567 & 0.568 & 0.153 & 1.36 & 1.45 & 2.05 \\
    \hline
    {} & score & 0.597 & 0.602 & 0.149 & 1.57 & 1.50 & 2.20 \\
    {} & consistency & 0.539 & 0.558 & 0.157 & 2.60 & 2.12 & 2.61 \\
    ite-2\hspace{-5pt} & merged & 0.573 & 0.587 & 0.153 & 2.00 & 1.80 & 2.51 \\
    \hline
    {} & score & 0.588 & 0.607 & 0.151 & 1.66 & 1.56 & 2.26 \\
    {} & consistency & 0.557 & 0.575 & 0.155 & 2.91 & 2.18 & 2.59 \\
    ite-3\hspace{-5pt} & merged & 0.582 & 0.592 & 0.151 & 2.18 & 1.93 & 2.58 \\
    \hline
    {} & score & 0.607 & 0.612 & 0.148 & 1.41 & 1.37 & 2.08 \\
    {} & consistency & 0.552 & 0.564 & 0.155 & 2.25 & 1.59 & 1.95 \\
    ite-4\hspace{-5pt} & merged & 0.578 & 0.590 & 0.152 & 2.16 & 1.92 & 2.52 \\
    \hline
  \end{tabular}
  }
  \end{minipage}
\end{table*}

\begin{figure*}
  \begin{minipage}[tl]{0.48\textwidth}
  \begin{center}
   \includegraphics[width=7.6cm]{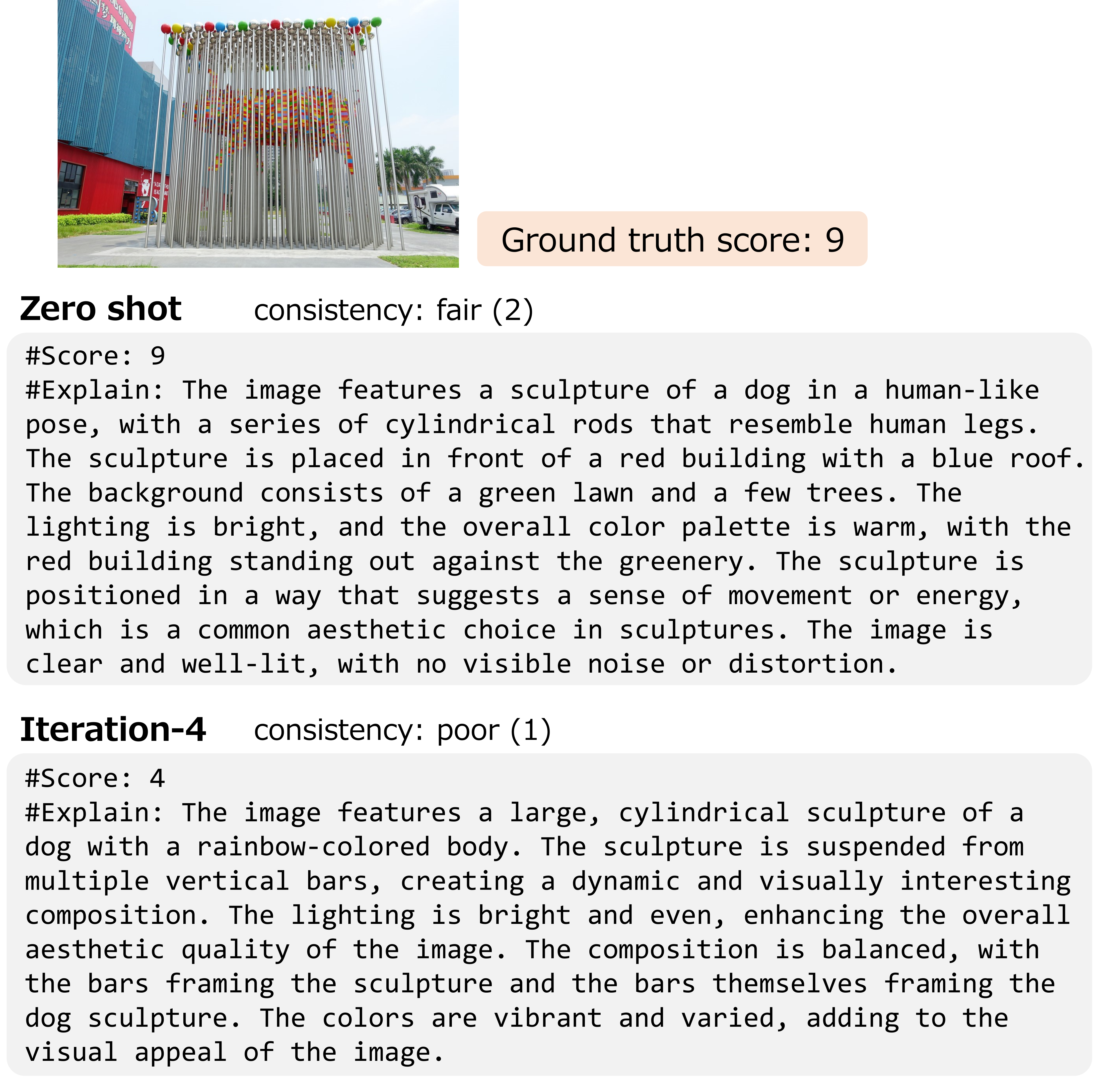} \\
   \vspace{5pt}
   \includegraphics[width=7.6cm]{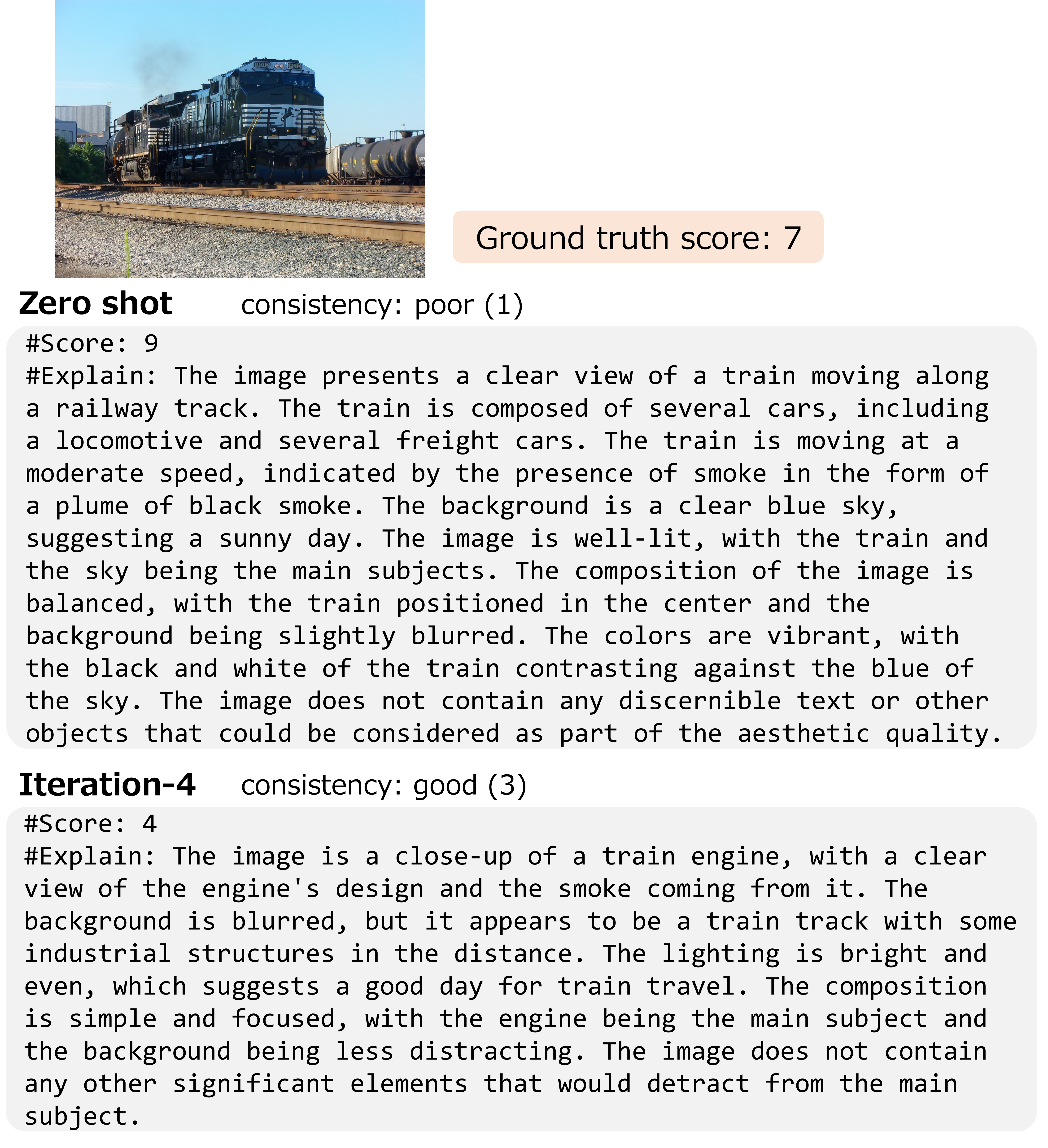}
  \end{center}
  \end{minipage}
  \begin{minipage}[tr]{0.48\textwidth}
  \begin{center}
   \includegraphics[width=7.6cm]{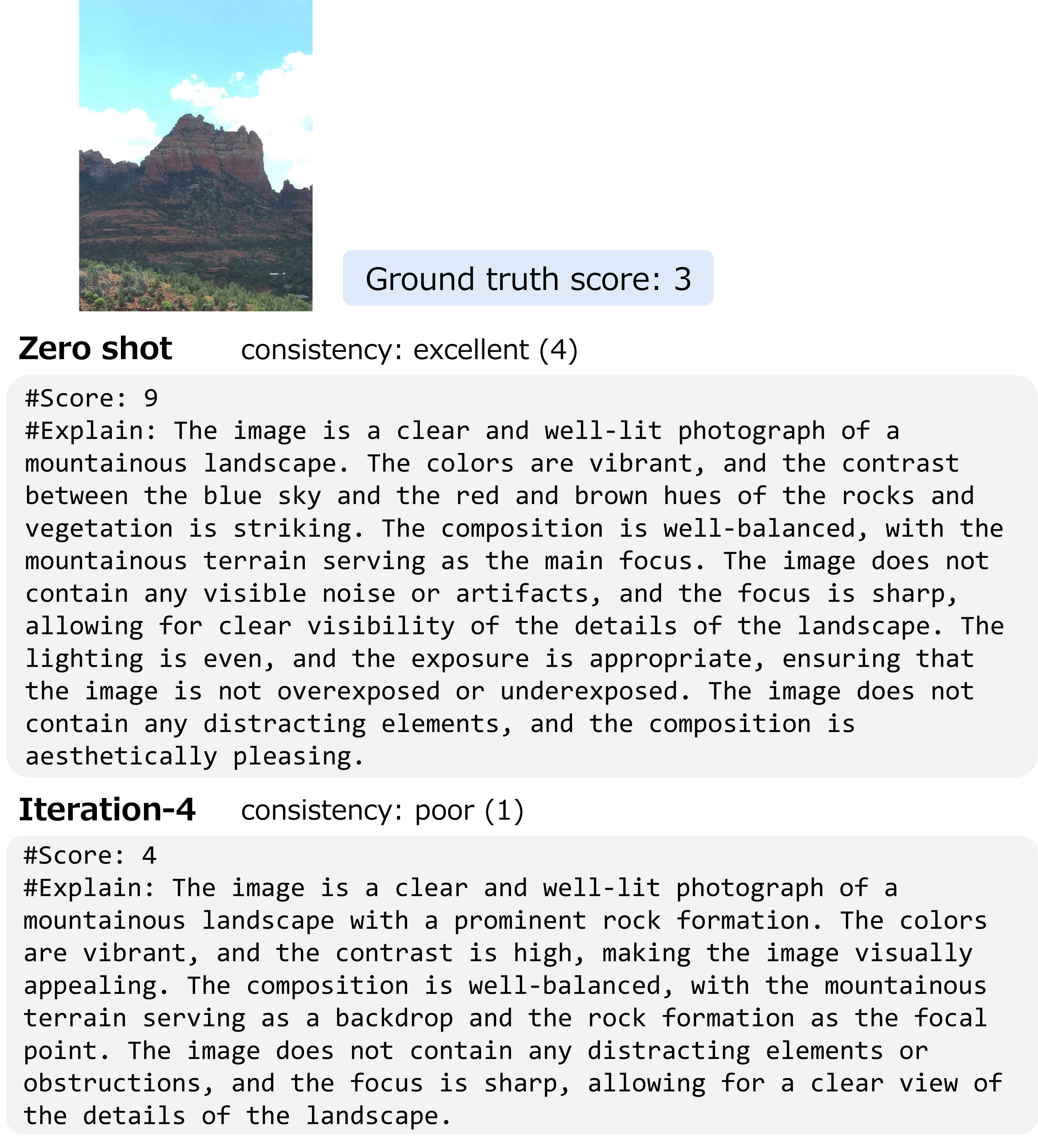} \\
   \vspace{5pt}
   \includegraphics[width=7.6cm]{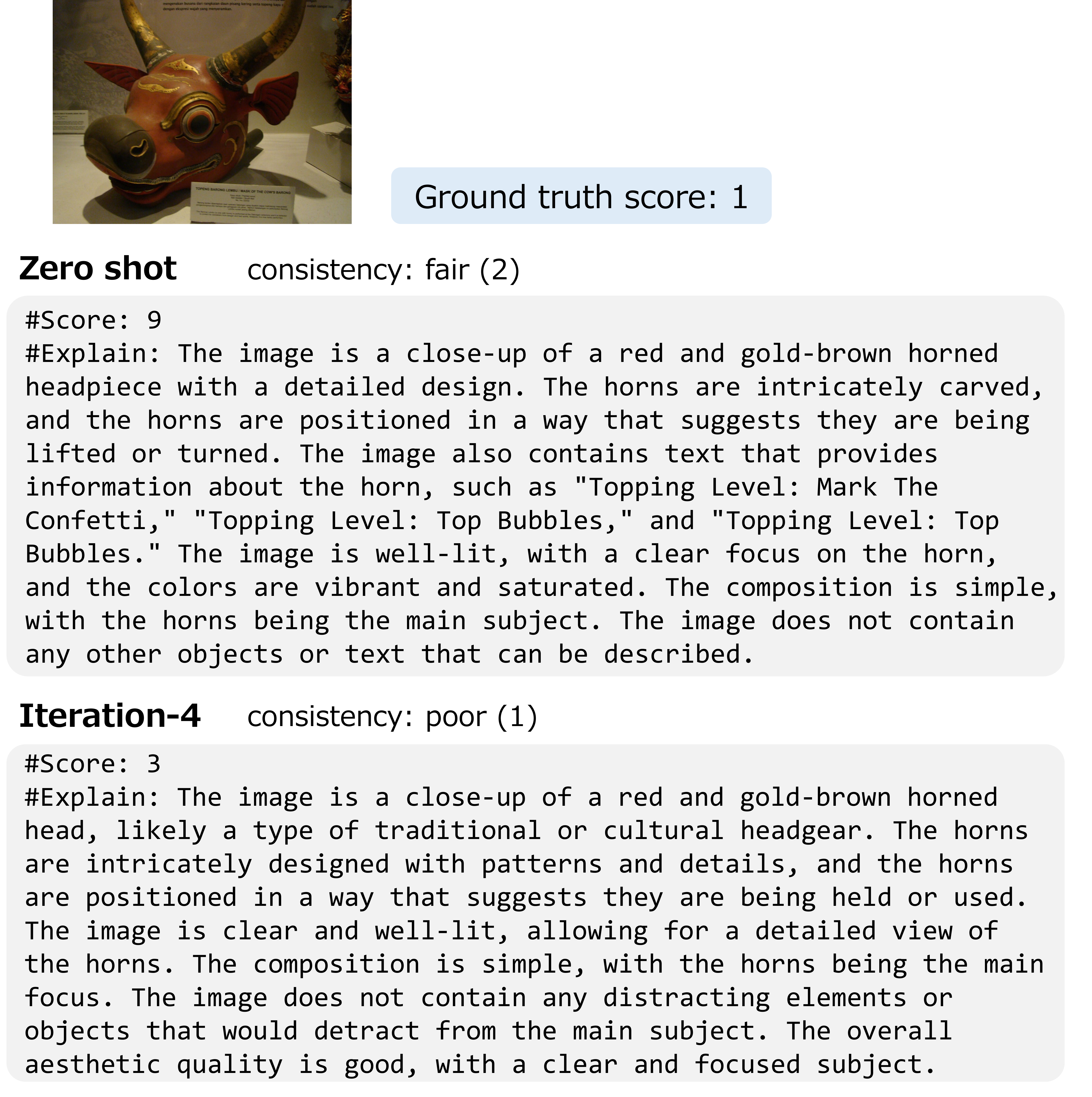}
  \end{center}
  \end{minipage}
 \vspace{-10pt}
 \caption{Output examples of the LLaVA-interleave-0.5B model on the AADB dataset.}
 \label{fig:example_interleave}
\end{figure*}

\section{Additional Results}
\label{sec:additional_results}
\subsection{Complete evaluation results}
\label{ssec:full_results}
In Tables \ref{table:full_next}-\ref{table:full_interleave}, complete evaluation results for all models are shown.
As LLM evaluation metrics, the usefulness metrics (\texttt{{\small Use}}) and the general writing quality (\texttt{{\small Gen}}) are presented in addition to the consistency metrics (\texttt{{\small Cons}}).

In Table~\ref{table:full_15}, the LLaVA-1.5-7B models trained on consistency data occasionally show notably low consistency scores.
This is primarily due to the model's outputs deviating from the required format, resulting in parsing failures and empty explanations being fed into GPT-4o. 
Interestingly, this issue is resolved after model merging.

\subsection{More example outputs}
\label{ssec:more_examples}
Figures~\ref{fig:example_next}-\ref{fig:example_interleave} present additional examples for model outputs sampled from the AADB test dataset for each model.

\subsection{Multi-turn dialogue}
\label{ssec:conversation}
\begin{figure}[t]
  \begin{center}
   \includegraphics[width=7.8cm]{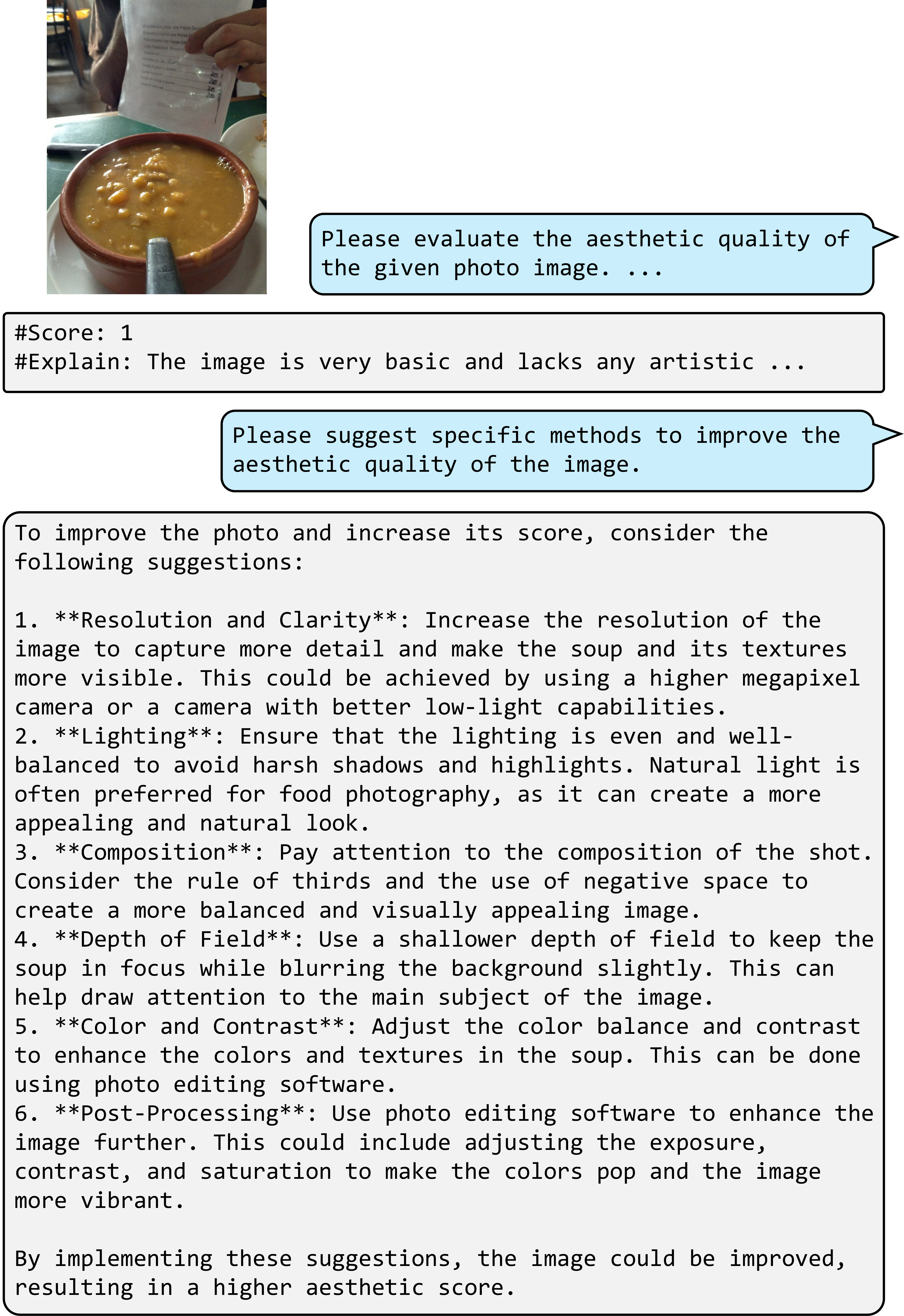}
  \end{center}
 \vspace{-20pt}
 \caption{Example of a multi-turn dialogue}
 \label{fig:conversation}
\end{figure}

When considering VLMs as AI assistants, the ability to engage in multi-turn dialogues is desirable, in addition to providing image scores and justifications.
Since our fine-tuning process uses a single prompt to elicit scores and justifications, the model's multi-turn dialogue capabilities might be compromised.
To assess these capabilities, we prompt the model with \textit{``Please provide specific suggestions for improving the photo to increase its score."} after it outputs the initial score and justification. 
This experiment uses the LLaVA-NeXT-7B model trained on the AADB dataset.
An example output is shown in Fig.~\ref{fig:conversation}.
The result indicates that the model can provide relevant responses even to questions not present in the fine-tuning data.

To quantitatively evaluate the results, we use GPT-4o to judge the usefulness, concreteness, and relevance of the generated responses with the following prompt:
\begin{coloredquotation}
\small
Please evaluate the AI assistant's response regarding the input image. The AI assistant received the instruction ``Please suggest specific methods to improve the aesthetic quality of the image" along with the image. Evaluate the response based on the following criteria: \\
usefulness: Whether the response is genuinely helpful in significantly improving the aesthetic quality of the image. For example, does it provide actionable advice that can lead to noticeable improvements? \\
concreteness: Whether the suggestions are highly specific and actionable. For example, does the response include detailed steps or techniques that can be directly applied? \\
relevance: Whether the suggestions are directly related to the content and context of the image. For example, are the suggestions tailored to the specific elements and style of the image? \\
The evaluation should be in five levels: bad, poor, fair, good, excellent. Output only in JSON format. \\
Example output: \{"usefulness": "poor", "concreteness": "good", "relevance": "fair"\} \\
The AI assistant's response is as follows: \\
\{text\}
\end{coloredquotation}
The images are also provided as input during the evaluation.
Table~\ref{table:conversation} presents the average scores across 500 samples from the test set.
The results demonstrate that our fine-tuned model maintains performance comparable to the base model on these metrics, indicating that the fine-tuning process does not degrade its dialogue capabilities.

\begin{table}[t]
  \caption{Evaluation of dialogue capabilities for the LLaVA-NeXT-7B model on the AADB dataset.}
  \label{table:conversation}
  \vspace{5pt}
  \centering
  \setlength{\tabcolsep}{3pt}
  {\small 
  \begin{tabular}{r|rrr}
    \hline
    {} & usefulness & concreteness & relevance \\
    \hline
    base model & 3.23 & 3.30 & 3.71 \\[-1pt]
    trained model & 3.22 & 3.28 & 3.66 \\[-1pt]
    \hline
  \end{tabular}
  }
\end{table}

\end{document}